\pdfoutput=1

\documentclass[11pt]{article}
\usepackage{enumitem}
\usepackage{style/eacl2023}
\usepackage{todonotes}

\usepackage{times}
\usepackage{latexsym}
\usepackage{graphicx}
\usepackage[T1]{fontenc}
\usepackage{bbold}
\usepackage{array}
\usepackage{mfirstuc}

\usepackage{tabularx}
\usepackage{array}
\usepackage{cellspace}
\newcolumntype{C}[1]{>{\centering\arraybackslash}m{#1}}

\usepackage[utf8]{inputenc}
\usepackage{tabularx}    
\usepackage{multirow}
\usepackage{microtype}

\usepackage{inconsolata}

\usepackage{subfiles} 

\usepackage[bb=ams]{mathalfa}

\title{

Improving the TENOR of Labeling: Re-evaluating Topic Models for Content Analysis
}

\author{
    Zongxia Li\textsuperscript{1} \hspace{2em} Andrew Mao\textsuperscript{1} \hspace{2em} Daniel Stephens\textsuperscript{3} \hspace{2em} Pranav Goel\textsuperscript{1} \\[0.2cm]
    \textbf{Emily Walpole}\textsuperscript{\textbf{2}} \hspace{2em} \textbf{Alden Dima}\textsuperscript{\textbf{2}} \hspace{2em} \textbf{Juan Fung}\textsuperscript{\textbf{2}} \hspace{2em} \textbf{Jordan Boyd-Graber}\textsuperscript{\textbf{1}} \\[0.2cm]
    \textsuperscript{1}University of Maryland, \texttt{\{zli12321, amao, pgoel\}@cs.umd.edu, jbg@umiacs.umd.edu} \\
    \textsuperscript{2}\abr{nist}, \texttt{\{emily.walpole, alden.dima, juan.fung\}@nist.gov} \\
    \textsuperscript{3}Morgan State University, \texttt{dstephens@morgan.edu} 
}

\newif\ifcomment\commenttrue

\usepackage[a-1b]{pdfx}

\usepackage{framed}
\usepackage{mdwlist}
\usepackage{siunitx}
\usepackage{latexsym}
\usepackage{colortbl}
\usepackage{xcolor}
\usepackage{nicefrac}
\usepackage{booktabs}
\usepackage{fnpct}
\usepackage{amsfonts}
\usepackage[T1]{fontenc}
\usepackage{bold-extra}
\usepackage{amsmath}
\usepackage{amssymb}
\usepackage{bm}
\usepackage{graphicx}
\usepackage{mathtools}
\usepackage{microtype}
\usepackage{multirow}
\usepackage{multicol}
\usepackage{xpatch}
\usepackage{latexsym,comment}
\usepackage[normalem]{ulem}

\newcommand*{\missingreference}{{\Huge \colorbox{red}{?reference?}}}
\newcommand*{\missingcitation}{{\Huge \colorbox{red}{?citation?}}}

\makeatletter
\xpatchcmd{\@setref}{\bfseries}{\missingreference}{}{}
\def\@citex[#1]#2{\leavevmode
    \let\@citea\@empty
    \@cite{\@for\@citeb:=#2\do
        {\@citea\def\@citea{,\penalty\@m\ }%
            \edef\@citeb{\expandafter\@firstofone\@citeb\@empty}%
            \if@filesw\immediate\write\@auxout{\string\citation{\@citeb}}\fi
            \@ifundefined{b@\@citeb}{\hbox{\reset@font\missingcitation}%
                \G@refundefinedtrue
                \@latex@warning
                {Citation `\@citeb' on page \thepage \space undefined}}%
            {\@cite@ofmt{\csname b@\@citeb\endcsname}}}}{#1}}
\makeatother

\newcommand{\gem}[1]{\mbox{\textsc{gem}}}
\newcommand{\abr}[1]{\textsc{#1}}

\newcommand{\lda}{\abr{lda}}
\newcommand{\slda}{{\small s}\abr{lda}}

\newcommand{\bertopic}{\abr{bert}{\small opic}}

\newcommand{\hidetext}[1]{}
\newcommand{\ignore}[1]{}

\ifcomment
    \newcommand{\pinaforecomment}[3]{\colorbox{#1}{\parbox{.8\linewidth}{#2: #3}}}

    \newcommand{\prtodo}[1]{\pinaforecomment{lightblue}{pr}{#1}}
    \newcommand{\prtodoi}[1]{\pinaforecomment{lightblue}{pr}{#1}}
\else
    \newcommand{\pinaforecomment}[3]{}
    \newcommand{\prtodo}[1]{}
    \newcommand{\prtodoi}[1]{}
\fi

\newcommand{\smallurl}[1]{ \begin{tiny}\url{#1}\end{tiny}}

\definecolor{lightblue}{HTML}{3cc7ea}
\definecolor{CUgold}{HTML}{CFB87C}
\definecolor{grey}{rgb}{0.95,0.95,0.95}
\definecolor{ceil}{rgb}{0.57, 0.63, 0.81}
\definecolor{UMDred}{HTML}{ed1c24}
\definecolor{UMDyellow}{HTML}{ffc20e}

\begin{document}

\maketitle
\begin{abstract}



Topic models are a popular tool for understanding text collections, but their evaluation has been a point of contention. Automated evaluation metrics such as coherence are often used, however, their validity has been questioned for neural topic models (\abr{ntm}s) and can overlook a model's benefits in real-world applications. To this end, we conduct the first evaluation of neural, supervised and classical topic models in an interactive task-based setting. We combine topic models with a classifier and test their ability to help humans 
conduct content analysis and document annotation.
From simulated, real user and expert pilot studies, the Contextual Neural Topic Model does the best on cluster evaluation metrics and human evaluations; however, \abr{lda} is competitive with two other \abr{ntm}s under our simulated experiment and user study results, contrary to what coherence scores suggest. We show that current automated metrics do not provide a complete picture of topic modeling capabilities, but the right choice of \abr{ntm}s can be better than classical models on practical tasks. 
\end{abstract}


\section{Introduction} \label{sec:sections/10-intro} 











Establishing a label set to organize a collection of documents is a
fundamental task in many fields such as social science, and, linguistics,
education. For example, in the social sciences, grounded
theory emphasizes \textit{structural coding} as a framework for
discovering similarities and differences in large-scale experimental
data and assigning meaning to it~\cite{glaser2017discovery,
Lindstedt2019, informatics8010019}.
Such a process is difficult and time-consuming, partly because it
requires a global understanding of the entire dataset, and local
knowledge to accurately label individual documents.
We emphasize that this strictly more general than document classification: classification presumes \textit{a priori} a label set; while we will use classifiers in our method, we first need a user's help to determine the label set and the training data.






Topic modeling~\cite{application_of_TM} has emerged as a popular tool to help with the
\textit{coding} process to discover the label set (Section~\ref{section:background}). 
These models treat documents as admixtures of latent topics, each represented by a distribution over words. 
The most popular topic model, Latent Dirichlet Allocation (\lda{})~\cite{blei2003lda} 
has over 40,000 citations with numerous extensions and variants~\cite{TMevolution}.

Previously, Active Learning with Topic
Overviews~\cite[\abr{alto}]{alto} demonstrated that
combining \lda{} with an active learning classifier could help
people create label sets more efficiently.  After topic models provide
a global overview of the data, exposing the broad themes of the
corpus, active learning selects documents that direct the annotator's
attention to topically distinct examples to label.
Together, these two ingredients train a classifier to automatically
label the documents more efficiently.

However, a gap remains in the literature, given recent advancements in
topic modeling.
Neural topic models (\abr{ntm}), which use continuous text embeddings
to capture contextual and semantic relationships in high-dimensional
data, have gained prominence, besting classical probabilistic topic
models on automatic evaluation metrics such as
coherence~\cite{aletras-stevenson-2013-evaluating}.
However, automated evaluation metrics have been called into question~\cite{li2024cfmatch}; 
\citet{neuralbroken} show they do not necessarily correlate with human ratings on topic model
outputs and call for task-centered evaluations, such as helping users analyze content.





We aim to fill this gap, and evaluate the effectiveness of neural,
supervised, and classical topic models to help social scientists with
content analysis and label set creation.
We do this by taking the starting point of \abr{alto}---classicial
topic models applied to this problem---and probe ``deeper'' to create
Topic-Enabled Neural Organization and Recommendations (\abr{tenor}),
an interactive tool that supports various topic models with active
learning to speed up the process of content
analysis.\footnote{\url{https://github.com/zli12321/TENOR.git}} We
conduct synthetic experiment on \lda{}, supervise \lda{} and
three \abr{ntm}s with followup user study and expert user study and
show that the choice of Contextualized Topic Model \cite{kitty_ctm}
(\abr{ctm}) helps users create higher quality label sets than using
classical \lda{}, as measured by both cluster metrics
(Section~\ref{section:evaluation_results}) and user ratings. However, \lda{} is
still competitive or better when compared with two other popular \abr{ntm}s. 
Thoughtfully using topic models as part of a larger system
with human interactions gives a more comprehensive evaluation and
understanding of their real-world usage (Section~\ref{user study}).

 \label{introduction}

\section{Background} \label{sec:sections/30-background} 







Manually sorting thousands of documents to establish a label set to create is mentally challenging and time-consuming. \citet{MimnoGrounded} compare grounded theory with topic modeling: although the two methods are from distinct fields, they produce similar insights on large-scale data. 
Topic models
cluster documents and extract meaningful themes and can help users induce labels. 

For content analysis, machine learning and \abr{nlp} focus on developing \abr{ntm}s~\cite{neuralbroken}, because they win nearly every automatic coherence metric. However, most of the computational social science community remains focused on older probabilistic models~\cite{ABDELRAZEK2023102131}.
Thus, we explore this open question: should we use classical or neural topic models for label induction and content analysis?






One of the reasons that \abr{ntm}s might be better is that \abr{alto}
showed the benefits of active learning~\cite{active-learning}: start with a
dataset with an undefined label set; users add labels to the set by going
through individual documents (guided by topic overviews); once the users
establish at least two distinct labels for the label set, a classifier trained
on the labeled documents can point users to documents that are either challenging for the current label set or that might require new labels.
One of the criticisms of \abr{ntm}s is that they are too granular and specific~\cite{neuralbroken}, but this may be a boon for label induction: it can find
candidates for a new label.

In addition to ignoring neural models (which had not reached maturity when \abr{alto} was proposed),
another lacuna of \citep{alto} is that it ignores supervised topic models that
can combine classification with topics. Supervised topic models~\cite{slda} \emph{change} as labels are added and can adapt---for
instance---when a user associates two labels with a topic. 
Thus we evaluate neural and classical topic models that tasks humans with creating a label set and annotating a document collection, with the assistance of topic models and a text classifier on the a dataset of \abr{us} congressional bills~\cite{AdlerWilkersonBillsProject}.

%
We delve into specific topic models, active learning, and evaluation metrics for the rest of this section.

\subsection{Topic Models}




Topic models identify latent themes within a corpus, providing a snapshot of
its overall narrative.
Given a set of documents and a specified topic count~$K$, these models divide documents into~$K$ clusters.
Each cluster represents
a topic defined by key terms, denoting its core theme (examples in Appendix~\ref{tab:topic_examples}).
Users can explore the corpus's main themes and label individual documents with the topics and keywords.




\paragraph{Supervised Latent Dirichlet Allocation.} 
\slda{} retains the generative process of \lda{} but also
adds a step to generate labels for each document \emph{given} its
empirical distribution over topic assignments in a document. For example, for movie comment reviews, \lda{}
generates general topics people discuss movies that are unlikely to correlated with users' star ratings.  In contrast, \slda{} can: an \lda{} topic about \underline{romance} films would split into ``good'' and ``bad'' versions with \slda{}.
%
We use the classifier's predictions as surrogate response
variables, and update \slda{} constantly as users label more
documents. 
We expect \slda{}'s topics to better reflect user inputs by interacting with the classifier trained with user
input labels.\footnote{Suppose a user creates 15 unique labels for 80 documents, we train the classifier on the 80 documents with the user input labels. Then we use the classifier to make predictions for all the documents and use the predictions as response variables for \slda{}}

\paragraph{Neural Topic Models} 
Current popular neural topic models include Contexualized topic models~\cite[\abr{ctm}]{kitty_ctm}, 
\bertopic{}~\cite{bertopicMark}, and Embedded topic model~\cite[\abr{etm}]{ETMtopic}. Theses neural models take advantage of pre-trained word embeddings with rich contextual information to enhance the quality of discovered topics. 
\abr{ctm} builds on pre-trained language models like \abr{sbert}~\cite{Reimers2019SentenceBERTSE} to generate sentence embeddings concatenated with Bag-of-Word (BoW) representations and runs a variational autoencoder (\abr{vae}) on the representation, while \bertopic{} uses \abr{umap}~\cite{mcinnes2020umap} and \abr{hdbscan}~\cite{McInnes2017hdbscanHD} create and refine topics from encoded word embeddings. \abr{etm} retains the same generative process as \lda{} but the topics are learned from word embeddings that contain rich semantic meanings instead of pure word distributions.


\subsection{Active Learning}

Active learning~\cite{active-learning} guides users' attention to examples
that would be the most beneficial to label for a classifier, using techniques
such as uncertainty sampling.  By directing users to annotate uncertain
documents first, active learning is valuable in situations constrained by time
or budget.


\subsection{Preference Functions}
During the initial stages of training, a classifier must generalize to unseen data quickly. 
A rapid improvement facilitates high-quality data analysis and optimizes time and costs, especially for large datasets~\cite{Muthukrishna_2019}.
%
Mathematically, ``preference functions'' are the tool that allows this early generalization in active learning generally and in \abr{tenor} specifically to get a good set of labels with representative documents as quickly as possible. 

A preference function uses uncertainty and diversity sampling to pick the most beneficial document and guide users' local attention to that document to label. 
%
According to the preference function, the classifier favors documents with the highest confusion scores that are most likely to be in the boundaries between multiple labels, which are documents that users are most likely to make new labels--uncertainty and diversity. For our baseline classifier, when it does not incorporate topic models, let $L$ be the label set probability distribution for document $d$, the preference function for $d$ is : 
%

\begin{equation}
\mathbb{H}_d(L) = -\sum_{i=1}^{n}{ P(l_i) \log P(l_i)}
.
\label{eq:baseline_preference}
\end{equation}

Here, \(\mathbb{H}_d \) represents the cross-entropy~\cite{Shannon1948} of the classifier. The ``most beneficial'' document is the one whose label distribution (as defined by a classifier) is most confused: more mathematically, has the highest entropy.  If the user can resolve that confusion by providing a new or existing label (or remove the document from the set), it will most benefit the next iteration of the classifier.

%

We follow the insight of \abr{alto} and interleave topic models and active learning to make the preference function topic-dependent.
This is important for real-world scenarios where context-switching can impede human labeling throughput~\cite{Raeburn2022ContextSwitching}.
%
First, the most confusing topic by the classifier is selected, and then, within this topic, the document with the highest preference function score (the most confusing document) is chosen.

Given $K$ topics from topic models, each document is characterized by a topic distribution vector \( \theta^d \equiv \{\theta^d_1, \theta^d_2, \ldots, \theta^d_K\} \). For a particular document, its predominant topic is: \begin{equation}
\theta^d_{\text{max}} = \max_{i=1}^{K} \theta^d_i.
\label{eq:topic_probability}
\end{equation}
We also adopt hierarchical sampling for active learning~\cite{flatTopic} and incorporates vector representation of topic models and users' label inputs to match their individual preferences~\cite{zhang2019sp10k}

\begin{equation}
\mathbb{H}_{d}^{t}(L) = \mathbb{H}_d(L) \cdot \theta^d_{\text{max}}.
\label{eq:topic_document_preference}
\end{equation}
With a clearly defined preference function, we choose a topic $k^*$ first based on the following criterion: Given \( K \) topics, let \( \mathcal{D}_k \) denote the set of all documents that are most prominently associated with topic \( k \). The classifier selects a topic \( k^* \) such that its documents' median preference score, \( \mathbb{H}_{d}^t \), is maximized. Formally, this is
\begin{equation}
    k^* = \underset{k \in \{1, 2, \ldots, K\}}{\arg\max} \text{median} \left\{ \mathbb{H}_{d}^t(L) : d \in \mathcal{D}_k \right\}.
\label{eq:topic_selection}
\end{equation}




\subsection{Evaluation Metrics}
\label{sec:metrics}
Our objective is for users to establish new label sets for a common
dataset.
This is a hard problem: indeed, \citet{NIPS2002_43e4e6a6} proves that it is impossible to satisfy multiple reasonable clustering properties simultaneously.
We thus, like \abr{alto} we use tree of reasonable metrics---described below---to compare how far user-induced labels deviate from a gold label set (in this case, the consensus labels of political scientists on the congressional bills dataset). 
In addition to these standard cluster evaluation metrics, we also measure the
coherence for each topic of \lda{}, s\lda{}, \abr{ntm}s (more detail in Appendix~\ref{sec:metrics_details}).


\paragraph{Purity} Purity evaluates how \emph{pure} an induced cluster is: in other words, what
proportion of documents in a cluster are not commingled with
documents with a different gold label~\cite{purity}. As we will see with many of
these metrics, there is a clear failure mode: the purity metric can
be easily manipulated by assigning a unique label to each document. We
mitigate this risk by not disclosing these metrics to labelers and limiting
the time users have to create labels.

\paragraph{Adjusted Normalized Mutual Information (\abr{anmi})} Normalized Mutual Information~\cite[\abr{nmi}]{NMI} assesses clustering quality by measuring the interdependence between true and predicted labels. One can gain insights of the true labels by understanding the predicted labels. The
\abr{anmi}~\cite{ANMI}, an enhancement of \abr{nmi}, corrects for the chance alignment of
clusters.

\paragraph{Adjusted Rand Index (\abr{ari})}
The Rand Index (RI)\cite[\abr{ri}]{RandIndex} measures for any pair of documents the probability that their gold labels and their assigned labels match.
The Adjusted Rand Index (ARI)~\cite[\abr{ari}]{ARI} refines this measure by adjusting for chance, which can yield negative values if the new labeling actively contradicts the gold labeling.

\paragraph{Coherence}
Normalized pointwise mutual information (\abr{npmi}) measures how semantically similar the top words of a topic are, which was proposed for classical topic models, but can also be used for
\abr{ntm}s~\cite{aletras-stevenson-2013-evaluating}.\footnote{\abr{npmi} and \abr{anmi} are over different evaluation metrics over different probability spaces.} \citep{NIPS2009_f92586a2} uses large-scale of user study to show coherence creates a computational proxy that simulates human judgments for classical topic models. We use \abr{npmi} to evaluate the quality of topics generated by classical and neural topic models.

The clustering metrics evaluate the alignment, quality, and information
overlap between two clusters. A higher value in these metrics indicates 
greater similarity and alignment between the induced labels and the gold labels. However,
using just one of them to measure user label quality has
limitations. If users assign every document a different label, they
will reach a perfect purity score, but that violates the task. \abr{ari} does
not measure the quality of individual clusters. For example, two clusters
might have high \abr{ari}, but both are very poor
quality. \abr{anmi} is sensitive to the number of clusters, where a
significant difference in the number of clusters between the standard cluster
and classifier predictions can lead to a reasonable \abr{anmi} score, but the
clusters have a high mismatch. By using all of them to complement each
other, we are more confident in comparing the quality of classifier
predictions.






 \label{section:background}

\section{Study Setup} \label{sec:sections/40-setup} \subsection{Groups}
\label{sec:groups}

For the simulated user study, we use the following models in combination with active learning: 
\begin{enumerate*}
    \item (\abr{none}); 
    \item Latent Dirichlet Allocation--(\lda ); 
    \item Supervised \abr{lda}--(\slda);
    \item Bertopic--(\bertopic{}); 
    \item Embedded Topic Model--(\abr{etm});
    \item Contextualized Topic Model--(\abr{ctm}).
\end{enumerate*}



Our baseline (1) \abr{none} gives users access to a classifier with active learning, but no topic model organization to help them first establish a ``big picture''. The rest of the groups provides the users topic overview and a classifier that has access to topic model probability vectors and active learning. More implementation details of our study groups are in Appendix~\ref{sec:study_group_details}.

\subsection{Dataset}
Our simulated experiment uses the 20newsgroups~\cite{misc_twenty_newsgroups_113} and the Congressional bills dataset. Both datasets have hierarchical labels; the first level is a general category, such as \textit{Health} or \textit{Education} for the Bills; and recreation (\textit{rec}) or science (\textit{sci}) for 20newsgroups. 
Under each of the first layer labels, there are more specific labels; for example, under \textit{Health}, there are \textit{Health Insurance}, \textit{Mental Health and Cognitive Capacities}, \textit{Children and Prenatal Care}, etc. 

Since we want to test our system theoretically and in a user study setting, having datasets with hierarchical labels enables us to use more specific labels as user input labels and more general labels as standard labels in simulated experiments. In real-world settings, users are more likely to make more specific labels that are more closely related to the contents of individual documents. 


\subsection{Simulated Experiment}

\begin{figure*}
    \centering
    \includegraphics[scale=0.8]{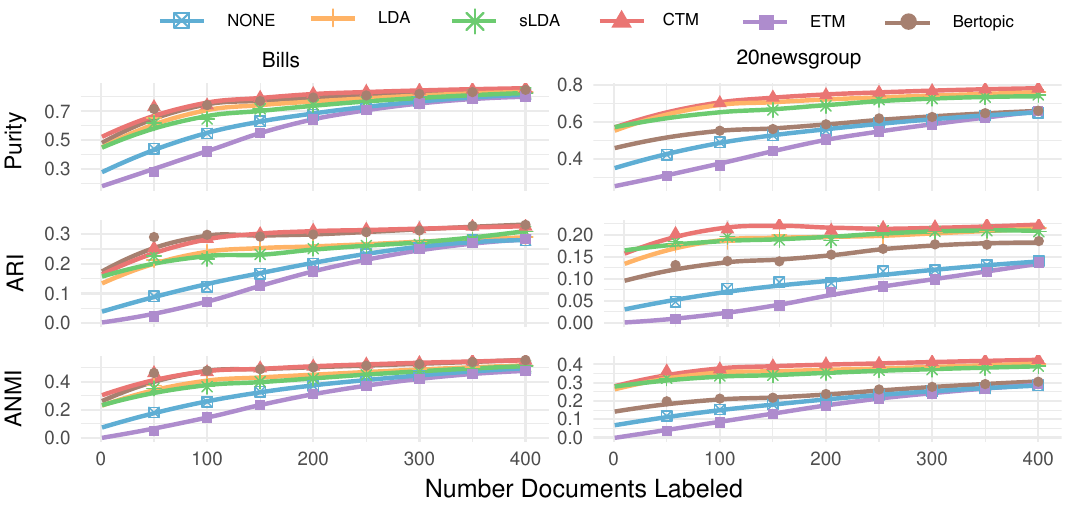}
    \caption{Cluster scores of simulated labeling experiments, median of 15 runs. \abr{ctm} with active learning has the highest score across all metrics and datasets.
    \abr{lda} and s\abr{lda} are better than or competitive with the other \abr{ntm}s (\abr{etm}, \bertopic{}).
    Given these results on synthetic data, we use \abr{ctm} for the human experiments.
    }
    \label{fig:synthetic}
\end{figure*}


Before conducting a real-world user study, we run simulated experiments on both datasets. 
We choose $K=35$ topics for all five topic models.\footnote{We choose $K=35$ because it optimizes average coherence for all topic models (details and hyperparameter selections are in Appendix~\ref{app:synthetic_experiment}).} 
Since users are more likely to create more detailed labels for each document, we use sub-labels as pseudo-user labels, while using the more general labels as our gold standard. 
We use logistic regression as our classifier and unigram tf-idf as input features for the classifier.\footnote{Using sentence transformer features produces similar results but takes much longer to update.} 
We also concatenate topic probability distributions for all the documents with tf-idf features, which encodes topic information to the classifier for settings with topic models. 
We use incremental leaning~\cite{incrementallearning} to fit and update the classifier after applying a synthetic label to each document.\footnote{With two exceptions\dots we  reinitialize the classifier: if a new label class is introduced to the classifier; if \abr{slda} is updated with surrogate response variables, we rebuild the features by concatenating tf-idf features with new topic information.} 
The clustering quality is assessed by the classifier's predictions with the more general labels using the three evaluation metrics. 
We run the experiment for 400 documents since we expect it to be the maximum for a participant to label within an hour. 

\paragraph{Coherence and simulated experiment results do not have a direct relationship} \abr{ctm} does the best on all cluster metrics on both datasets (Figure~\ref{fig:synthetic}), while \abr{lda} and s\abr{lda} remain competitive with other \abr{ntm}s.
Topic models with higher \abr{npmi} in Table~\ref{tab:coherence} do not necessarily have better simulated experiment results shown in Figure~\ref{fig:synthetic}. \abr{etm} does the worst among all the groups---despite having high coherence---and \abr{ctm} does the best, where \abr{lda} and s\abr{lda} are even better than \bertopic{} and \abr{etm} on the 20newsgroup dataset.

\begin{table}[t]
\small
\centering
\begin{tabular}{cccccc}
\hline
\rowcolor{gray!50}
Dataset & \abr{lda} & s\abr{LDA} & \abr{ctm} & \abr{etm} & \bertopic{} \\ \hline
Bills & 0.07 & 0.09 & 0.09 & 0.13 & 0.15 \\ 
Newsgroup & 0.06 & 0.05 & -0.09 & 0.09 & 0.10 \\ \hline
\end{tabular}
\caption{On average, \abr{ntm}s have higher \abr{npmi} coherence than \abr{lda}, where \bertopic{} has the highest coherence, followed by \abr{etm}. However, the \abr{ntm}s with higher coherence are not better than \abr{ctm} and \abr{lda} under a task-based experiment (Figure~\ref{fig:synthetic}).}
\label{tab:coherence}
\end{table}

While our synthetic data can serve as partial proxy, relying solely on automated evaluation metrics does not capture how much the users find the topic model helpful in helping them conduct content analysis. 
Thus, our next section investigates this question and surveys users' ratings on how they find topic models useful.

 \label{section:setup}


\section{User Study} \label{sec:sections/60-study} 


We conduct a general user study and expert study to compare topic models in the rest of the paper.
%
For the general user study, we compare settings (1) \abr{none}, (2) \lda{}, (3) \slda{}, and (6) \abr{ctm} since \abr{ctm} is the best-performing neural model in the simulated experiment.
We use the Bills dataset to conduct a 60-user study with our interface, with 15 people in each group. 
Our Bills dataset's topics are accessible to lay annotators and allows us to quantitatively understand users' acclimation to the dataset as they explore the corpus.
We then run a smaller, more qualitative followup expert study on an expert dataset, where the experts are familiar with the topics in the dataset with the best two models from our user study results. 
The goal of the expert study is to ensure that our user study results can generalize to experts with deeper knowledge of \abr{us} federal policy.  


\subsection{User Study Interface}
\label{sec:interface_setup}

For the groups using topic models, users are shown documents grouped by their top topic, with topic keywords.
The document selected by the active learning preference function is highlighted and displayed both at the top of its topic and at the top of the interface. 
When users click a document, they are presented with its full text, label options, top five topics, and top ten keywords per topic. 
Words above a 0.05 threshold in the primary topic are highlighted. 
In \abr{none} group, users see unsorted documents with the recommended one at the top.
Clicking a document shows its contents, without topic keywords or highlights (detailed interface in Appendix~\ref{app:user_interface}).


\subsection{Participant Recruitment}
We sourced participants via Prolific, restricting our selection to individuals from the US with an approval rate exceeding \(95\%\) with at least ten previous participations on Prolific. 
Participants were randomly assigned to one of four groups, each accommodating a maximum of fifteen participants.\footnote{We use the same trained models from the simulated experiment.
We update \slda{} in the backend once the previous training is complete.} Participants first reviewed the task instructions and completed a brief tutorial to familiarize themselves with the process. 
Participants complete a follow-up survey to receive a 20-dollar compensation after the one-hour session. 



\begin{figure}
    \centering
    \hspace*{-0.55cm}
    \includegraphics[scale=0.55]{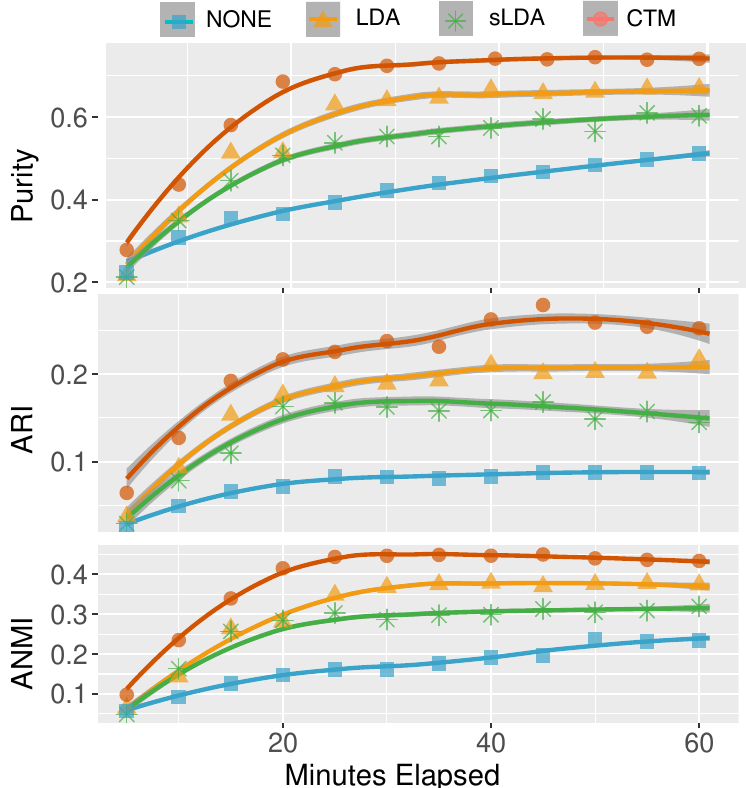}
    \caption{
    User study label cluster metrics plotted against time.
    For each group, we take the median of each metric for every minute passed. The user study results are similar to the simulated experiment; \abr{ctm} does the best on all three clustering metrics.}

    \label{fig:time_metrics}
\end{figure}

\subsection{Cluster Quality Evaluation Metrics}
\label{section:evaluation_results}

We record the purity, \abr{ari}, and \abr{anmi} for every minute passed during each session. 
For each group, we plot the median of each metric for every minute passed (Figure~\ref{fig:time_metrics}).

\paragraph{Topic model groups do better than \abr{none}} Throughout the 60-minute study session, the classifier has a wide gap between groups with topic models and \abr{none}. Topic model groups have faster early gains on all three metrics than \abr{none}, confirming the results from \citet{alto}.

\begin{figure*}[t]
    \centering
    \includegraphics[scale=0.55]{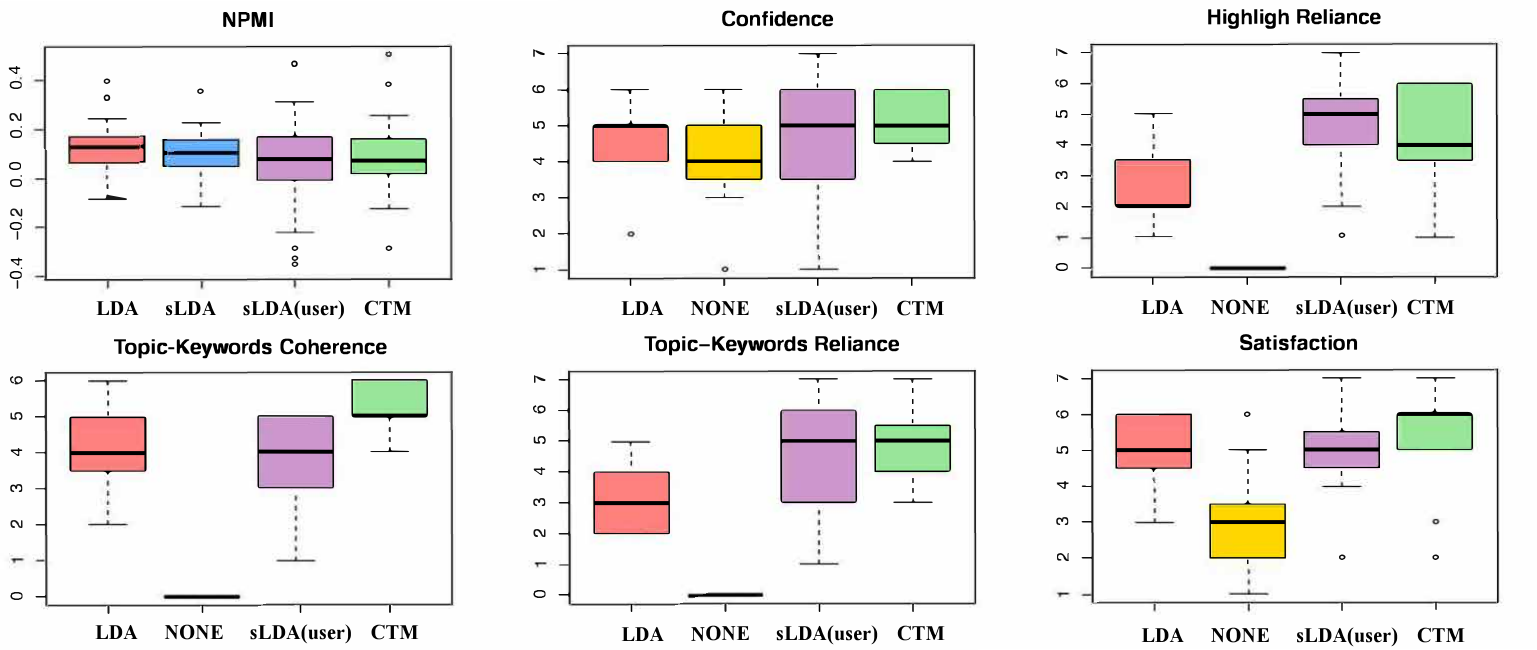}
    \caption{The first Plot shows \abr{npmi} Coherence for all topics on the Bills dataset, where \slda{}(user) is trained on user input labels, and \slda{} is the initial model used for all \slda{} users.
The rest of the plots shows users' rating on different questions on a scale 1 to 7, which the higher is better. Although \slda{} is worse than \lda{} and \abr{ctm} on clustering evaluations, most of the median of user ratings do not differ from \abr{ctm}, and surpass \lda{} in some ratings. For ratings 2 to 4, \abr{none} groups users all rate 0 because they do not have access to those features}
    \label{fig:user_rating1}
\end{figure*}

\paragraph{\abr{ctm} does the best on cluster metrics, followed by \abr{lda}, s\abr{lda}, and \abr{none}.} 
In real-world user applications, \abr{ctm} is the best for classification. 
The classifier with neural topic features, trained on user input labels, can generalize unseen data better than classical generative topic probability features. 
Although \abr{ctm} is the best, having the classifier have access to topic model features is better for the classifier to generalize and predict unseen data than not.
%
We later manually evaluate the validity of the user labels by random sampling (Appendix~\ref{app:sanity_check}), where 98.38\% of the selected examples are qualified under evaluations of two authors.



\paragraph{\slda{} falters on compared to \lda{} and \abr{ctm}} 
This is partly attributed to inaccuracies in the classifier's predictions.
For instance, when a user labels 30 documents midway through the session, the classifier, in turn, predicts labels for the entire dataset. 
However, if the user only creates two label categories for the 30 documents, the lack of diversity of response variables can generate document topic probability as features that confuses the classifier.
%
Nonetheless, \slda{} can align certain topics with user intent labels, which means that \slda{} might be capable of generating topic keywords that are semantically similar to user labels, thus improving users' overall experience. Subsequent survey analyses will investigate whether \slda{} supports this hypothesis in user survey ratings.



\paragraph{Examining coherence, quality of document clusters, and quality of topic keywords} 
We go through the topics with top two, middle two, and bottom two coherence scores for the models we use for user study (including s\abr{lda} trained on user labels), and show the \abr{npmi}, topic keywords, and a randomly selected passage from the topic in Tables~\ref{tab:topic_examples} and~\ref{tab:low_topic_examples}.\footnote{We load the saved \slda{} model trained on user labels predicted by the classifier at the end of the session, we call it \slda{}(user).}
Although the coherence scores vary for different topics, the top keywords are representative of the documents, but a low median coherence score does not necessarily show lower median user ratings (Figure~\ref{fig:user_rating1}). \abr{ctm} has the highest top coherence scores but the median coherence score is lower than \slda{} and \lda{}. However, \abr{ctm} is still better on clustering evaluations and user ratings.

\subsection{User Ratings}
\label{section:user_ratings_analysis}
Our survey comprises five questions aimed at gauging user judgment and evaluating topic models, using a scale ranging from 1 to 7. 
\footnote{\textbf{Confidence} asks how confident the users feel about their created labels.
\textbf{Highlight Reliance} asks how much the users rely on the \textit{highlight} functionality to make labels. 
\textbf{Topic-Keywords Coherence} asks whether users find the topic keywords coherent while they explore topics and peruse keywords to assist them in label creation. 
\textbf{Topic-keyword Dependence} investigates the frequency at which users consult the most related topic keywords while creating labels for documents. 
\textbf{Satisfaction} assesses the users’ overall satisfaction with the tool, exploring whether users find the tool likable and helpful.}


\paragraph{\abr{ctm} and \slda{} users rely more on topic models than \lda{}} 
Figure~\ref{fig:user_rating1}, the second to sixth plot show a summary of users' ratings for question 1 to 5.
The median of user ratings on \abr{ctm} and \slda{} are similar for most of the questions except for \textbf{Topic-Keyword Coherence}, which \slda{} falls short.
Based on the median user ratings, users generally rely more on topic keywords and highlights to create labels for documents if they are assigned to the \abr{ctm} or \slda{} group. 
Users also rate the topic keywords they use to label documents as more coherent for \abr{ctm} and \slda{}.
Although the classifier in \slda{} falls short on the three cluster metrics among the three topic models, users generally have better overall experience with \slda{} than \abr{lda} users.


\paragraph{Automatic coherence likes \lda{} topics, users do not}  Although the top topics for \abr{ctm}, \slda{} and s\abr{lda}(user) have higher coherence scores than \lda{} (Figure~\ref{fig:user_rating1}), \lda{}'s coherence scores are quite tight in the boxplot and \lda{} has higher median coherence than the other two models. 
\slda{}(user) has diverse coherence scores for its topics. 
However, when looking at the median user rating of all the five questions, \lda{} does not surpass \abr{ctm} and s\abr{lda}: there is not a strong and direct relationship between coherence and human usability.
This is a task-specific confirmation of \citet{hoyle2021automated}.


\begin{figure}[t]
    \centering
    \hspace*{-0.55cm}
    \includegraphics[scale=0.5]{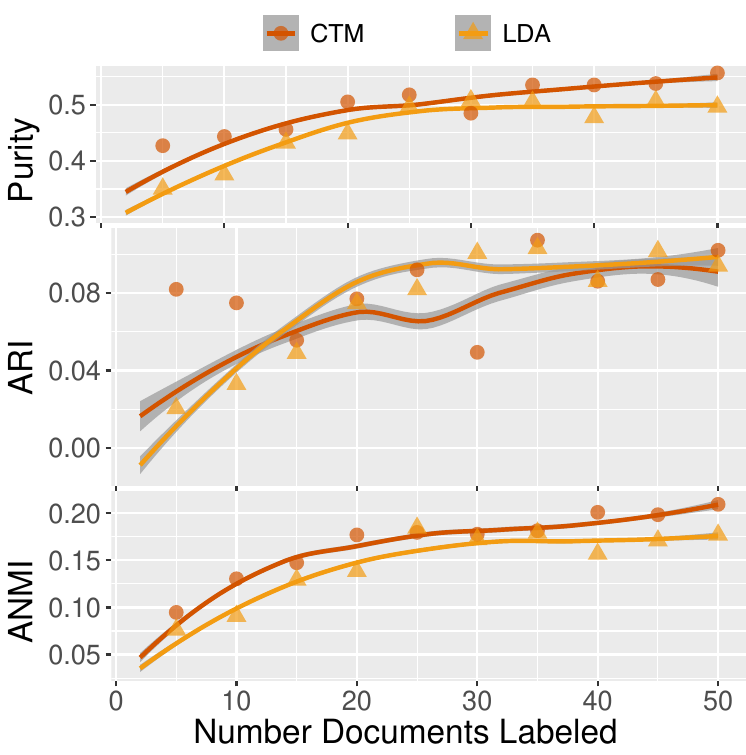}
    \caption{We run a followup pilot study with six social science experts (three in each group) on their internal social science dataset (800 documents). They are familiar with the topics in the dataset. Up to the 50th document labeled, \abr{ctm} still generalizes well for expert datasets and expert users.}
    \label{fig:user_rating1}
\end{figure}

\paragraph{Different topic models, different purposes} We run ANOVA~\cite{ANOVA} and posthoc Turkey-Kramer for pairwise comparison between ratings of any two of the user groups. 
Users are less likely to rely on the topic keywords generated by \lda{} to label documents, compared to \abr{ctm} and \slda{} based on significance results (Table~\ref{tab:significance test}) because \lda{} generates overly general topic keywords that are less useful to label individual documents.
For specific tasks, such as label set establishment and tasks involving understanding individual documents, \abr{ctm} is a better choice.

\begin{table}[t]
\small
\centering
\begin{tabular}{lcccl}
\toprule
\textbf{Metric} & \textbf{p-Value} & \textbf{Significant Pair} \\
\midrule
\midrule
Confidence & 0.327 & None \\
HighlightReliance & 0.035 & s\abr{lda} vs. \abr{lda} \\
topicCoherence & 0.017 & \abr{ctm} vs. s\abr{lda} \\
topicReliance & 0.034 & \abr{ctm} vs. \abr{lda} \\
satisfaction & 0.002 & \abr{none} vs. Other 3 \\
\bottomrule
\end{tabular}

\caption{Significance test results across subjective ratings for three groups at a 0.05 significance level. There is no significant difference in user ratings between \abr{ntm} and \abr{lda} except for \textit{Topic-Keyword Reliance}. For rows 2-4, we exclude \abr{none} to do testing. The third column shows the group pairs that are statistically significant. For example, the significant pair for \textit{satisfaction} is between \abr{none} and other three groups with topic models, and it indicates a difference of \textit{user satisfaction} rating between \abr{none} and other three groups under a 95\% confidence level, where the \abr{none} users are less satisfied with their experience from the sixth plot in Figure~\ref{fig:user_rating1}.}
\label{tab:significance test}
\end{table}

\subsection{Expert Verification}
The expert conditions were \lda{} and \abr{ctm}, the two winning conditions in our general user study.
Six experts all hold at least a graduate degree in community resilience related field that focuses on assisting communities and stakeholders on issues related to anticipated hazards conditions and disaster preparedness field.\footnote{\url{https://www.nist.gov/community-resilience}} 
We use the same user interface described in Section~\ref{sec:interface_setup} with the given expert dataset on 800 documents. 
%
The documents are collected from local governments across the United States providing structured ways to set community-scale goals and developing plans for recovery of community functions after natural or human-caused hazards~\cite{NISTSP1190GB16}. Experts conduct analysis and assign labels to this dataset so they can understand different categories of hazards and develop plans for a community to prepare for anticipated hazards, adapt to changing conditions, and withstand and recover rapidly from disruptions. The dataset has been previously labeled by multiple experts using Cohen's Kappa agreement~\cite{McHugh2012} over a six-month period. 
%
\abr{ctm} surpasses \lda{} on two out of three clustering metrics and has similar \abr{ari} at the 50\textsuperscript{th} document (Figure~\ref{fig:user_rating1}). 

\paragraph{Experts rely less on keywords but still like them} 
Since all the experts are quite familiar with the topics in the dataset, one expert using \lda{} mentions that the topic keywords are not helpful but the highlighted texts are more helpful for individual document annotation.
\lda{} produces topics that are too general, so experts already prefer the more specific keywords from \abr{ctm}.

 \label{user study}


\section{Related Work} \label{sec:sections/70-related} 



Applications of topic models are important, as exemplified by previous work by Fang et al~\cite{fang2023usercentered}, which addresses the human-centric applications of topic models. 
\citet{ITM-1} shows that interactive topic models have gained traction among social science researchers and data analysts. 
Nevertheless, classical topic models dominate most applications in social science research~\cite{application_of_TM, lin_bmcbioinformatics2009}. 
Despite their theoretical advantages, this persistent preference for classical models underscores the need for comprehensive studies on the practical utility of \abr{ntm}s.

As one of the most popular topic models, \lda{} has been widely applied and tested in diverse fields from health~\cite{Paul2011YouAW} to political opinion analysis~\cite{TMdatamining},  social media data analysis~\cite{Zhao2011ComparingTA}, etc.\ 
Thus, \lda{} has already proved itself as a useful tool for real applications. 

For supervised models, most work focuses on \slda{}'s power to predict response variables from text~\cite{xu2022supervised}.
Few works have study whether the induced topics align with user intents such as labeling.
Using \slda{} interactively for document recommendation and annotation is more intuitive and straightforward than using unsupervised classical \lda{}. 


Beyond connecting a single response variable to topic assignments, neural models offer even more flexibility and have over a hundred variants, but the evaluation of \abr{ntm}s is mainly based on topic coherence, topic diversity, and classification applications~\cite{ijcai2021p638}. 
The major framework of \abr{ntm}s are mostly sequential \abr{ntm}s, which leverages the architecture power of Recurrent Neural Network (\abr{rnn}); \abr{ntm}s with pre-trained language models, such as \abr{bert}, that already learns the semantic relationship and association of words from a large corpus of texts. 
\abr{ntm}s have the advantage of producing higher automatic evaluation scores, and classification abilities, along with other more extensive applications that classical topic models cannot do, which includes texts generation~\cite{tang-etal-2019-topic, wang-etal-2019-topic}, summarization~\cite{pmlr-v108-zhao20c, wang-etal-2020-friendly}. 

However, with the new popularity of \abr{ntm}s,  to the best of our knowledge, there are still few works using \abr{ntm}s for social science due to their complex architecture and more computing resource demands.
Our work examines this gap to study the trade-off between using neural, supervised, or classical topic models. 
While some recent studies have compares \abr{ntm} and \lda{} with human analysis of the topic outputs, they still predominantly rely on automatic evaluation metrics, with limited emphasis on analyzing the quality of models from a human perspective or task-based utility of topic models~\cite{BenchmarkingNT}. 
%
\citet{ItalianCaseStudy} concludes that \lda{} is better than \abr{ntm} in metrics on coherence\cite{coherence} and classification~\cite{classification}. 
However, this conclusion is for non-English datasets. Our research intends to bridge this gap by conducting an English-language topic model quality evaluation, incorporating human interaction to help content analysis.

Our approach differs from previous studies, which compares \abr{ntm}s and classical models' stability and alignment with stationary, pre-determined ground truth labels~\cite{hoyle2021automated}. 
In the former, \lda{} was better; in the latter, \lda{} was better than many \abr{ntm}s~\cite{neuralbroken}. However, \citet{hoyle2021automated}'s approaches only evaluate topic models by analyzing human ratings on topic keywords with labels without any task applications.
In contrast, for the tasks of content analysis and building a label set, the overly specific \abr{ntm} keywords are actually helpful for people to come up with labels more easily than more general and dispersed keywords. While the overall topics may not look as ``pretty'' to a user, they are useful.

\label{related work}

\section{Conclusion} \label{sec:sections/90-conclusion} 
We provide an interactive task-based evaluation of neural, supervised, and classical topic models, using the task of content analysis and label set creation. 
Using \abr{ctm} with an active learning classifier helps both expert and non-expert annotators produce higher quality label sets more quickly, according to cluster metrics and human ratings, validating that the right choice of \abr{ntm}s can be better than \abr{lda} for content analysis. However, \abr{lda} is still competitive with two other \abr{ntm}s, contrary to what coherence scores would suggest. We show that current automated metrics do not provide a complete picture of topic modeling capabilities, but the right choice of \abr{ntm}s can still be better than classical models on practical tasks. With the popularity of large language models (\abr{llm}s), future work can include exploring more effective ways to use \abr{tenor} combined with \abr{llm}s for content analysis, where experts have a set of pre-defined research question and hypothesis, and use \abr{tenor} to actively select documents to prompt an \abr{llm} to build up a label set for the dataset quickly to answer their research questions and verify their research hypothesis. 

 \label{conclusion}

\section{Limitations} \label{sec:sections/110-limitations} We provide a human-in-the-loop framework to evaluate topic models, extending beyond automated evaluation metrics. Yet, our experiment only focuses on a very narrow and specific task to evaluate topic models. In addition, although our work shows that the right choice of \abr{ntm} can bee more powerful than \abr{lda} for specific tasks, the debate about evaluation of topic models is still present. From a language perspective, our experiments are based on English dataset only. Our conclusions theoretically can be generalized to some other languages but need to be practically tested. It might come to a different conclusion for languages with completely different structures than English. Furthermore, with the rise of \abr{llm}s that can complete various tasks close to human level, the use of \abr{llm} to help with the process of label set generation, classification~\cite{zhou2024explore}, and content analysis is a more efficient and cost-effective approach that can simulates our human study compared with our human study in Section~\ref{user study}. A comparative analysis of the quality of labels created by actual human users and \abr{llm} would be valuable for the social science and \abr{nlp} community to confirm the validity of using \abr{llm}s to simulate actual user studies to speed up their research process. We will conduct further comparative analysis between human created labels and \abr{llm} created labels in our future work.





 \label{limitation}

\section{Ethics} \label{sec:sections/120-ethics} We received approval from the Institutional Review Board before initiating the user study. All participants are based in the United States. Users are required to review an instruction and consent statement before participation commitment. They have the option to withdraw if they disagree with the terms. Throughout the study, no personal information that could reveal identities is collected. To the best of our knowledge, our study presents no known risks. \label{ethics}

\section{Acknowledgement} \label{sec:sections/130-acknowledgements} We thank anonymous reviewers and Alexander Hoyle and Kyle Seelman for their insightful comments for helping us make our paper experiments and arguments more solid. We thank Emily Walpole and Juan Fung's community resilience groups for taking their time participating our expert verification experiment and providing valuable qualitative comments and feedback. Zongxia Li, Andrew Mao, Daniel Stephens, and Pranav Goel's contributions are supported by the \abr{nist} Professional Research Experience Program.  \label{acknowledge}

\bibliography{bib/custom}

\begin{thebibliography}{59}
\expandafter\ifx\csname natexlab\endcsname\relax\def\natexlab#1{#1}\fi

\bibitem[{Abdelrazek et~al.(2023)Abdelrazek, Eid, Gawish, Medhat, and Hassan}]{ABDELRAZEK2023102131}
Aly Abdelrazek, Yomna Eid, Eman Gawish, Walaa Medhat, and Ahmed Hassan. 2023.
\newblock \href {https://doi.org/https://doi.org/10.1016/j.is.2022.102131} {Topic modeling algorithms and applications: A survey}.
\newblock \emph{Information Systems}, 112:102131.

\bibitem[{Adler and Wilkerson(2008)}]{AdlerWilkersonBillsProject}
E.~Scott Adler and John Wilkerson. 2008.
\newblock {Congressional Bills Project}.
\newblock \url{http://www.congressionalbills.org}.
\newblock Accessed: insert access date here.

\bibitem[{Aletras and Stevenson(2013)}]{aletras-stevenson-2013-evaluating}
Nikolaos Aletras and Mark Stevenson. 2013.
\newblock \href {https://aclanthology.org/W13-0102} {Evaluating topic coherence using distributional semantics}.
\newblock In \emph{Proceedings of the 10th International Conference on Computational Semantics ({IWCS} 2013) {--} Long Papers}, pages 13--22, Potsdam, Germany. Association for Computational Linguistics.

\bibitem[{Amelio and Pizzuti(2016)}]{ANMI}
Alessia Amelio and Clara Pizzuti. 2016.
\newblock \href {https://doi.org/10.1111/coin.12100} {Correction for closeness: Adjusting normalized mutual information measure for clustering comparison: Correction for closeness: Adjusting nmi}.
\newblock \emph{Computational Intelligence}, 33.

\bibitem[{Bakharia et~al.(2016)Bakharia, Bruza, Watters, Narayan, and Sitbon}]{ITM-1}
Aneesha Bakharia, Peter Bruza, Jim Watters, Bhuva Narayan, and Laurianne Sitbon. 2016.
\newblock \href {https://doi.org/10.1145/2854946.2854960} {Interactive topic modeling for aiding qualitative content analysis}.
\newblock In \emph{Proceedings of the 2016 ACM on Conference on Human Information Interaction and Retrieval}, CHIIR '16, page 213–222, New York, NY, USA. Association for Computing Machinery.

\bibitem[{Baumer et~al.(2017)Baumer, Mimno, Guha, Quan, and Gay}]{MimnoGrounded}
Eric P.~S. Baumer, David Mimno, Shion Guha, Emily Quan, and Geri~K. Gay. 2017.
\newblock \href {https://doi.org/10.1002/asi.23786} {Comparing grounded theory and topic modeling: Extreme divergence or unlikely convergence?}
\newblock \emph{J. Assoc. Inf. Sci. Technol.}, 68(6):1397–1410.

\bibitem[{Bianchi et~al.(2021)Bianchi, Terragni, and Hovy}]{kitty_ctm}
Federico Bianchi, Silvia Terragni, and Dirk Hovy. 2021.
\newblock \href {https://doi.org/10.18653/v1/2021.acl-short.96} {Pre-training is a hot topic: Contextualized document embeddings improve topic coherence}.
\newblock In \emph{Proceedings of the 59th Annual Meeting of the Association for Computational Linguistics and the 11th International Joint Conference on Natural Language Processing (Volume 2: Short Papers)}, pages 759--766, Online. Association for Computational Linguistics.

\bibitem[{Blei et~al.(2003)Blei, Ng, and Jordan}]{blei2003lda}
David~M Blei, Andrew~Y Ng, and Michael~I Jordan. 2003.
\newblock Latent dirichlet allocation.
\newblock \emph{Journal of machine Learning research}, 3(Jan):993--1022.

\bibitem[{Boyd-Graber et~al.(2017)Boyd-Graber, Hu, and Minmo}]{application_of_TM}
Jordan Boyd-Graber, Yuening Hu, and David Minmo. 2017.
\newblock \href {https://doi.org/10.1561/1500000030} {\emph{Applications of Topic Models}}.
\newblock Now Foundations and Trends.

\bibitem[{Chang et~al.(2009)Chang, Gerrish, Wang, Boyd-graber, and Blei}]{NIPS2009_f92586a2}
Jonathan Chang, Sean Gerrish, Chong Wang, Jordan Boyd-graber, and David Blei. 2009.
\newblock \href {https://proceedings.neurips.cc/paper_files/paper/2009/file/f92586a25bb3145facd64ab20fd554ff-Paper.pdf} {Reading tea leaves: How humans interpret topic models}.
\newblock In \emph{Advances in Neural Information Processing Systems}, volume~22. Curran Associates, Inc.

\bibitem[{Chen et~al.(2010)Chen, Zhu, Kifer, and Lee}]{TMdatamining}
Bi~Chen, Leilei Zhu, Daniel Kifer, and Dongwon Lee. 2010.
\newblock What is an opinion about? exploring political standpoints using opinion scoring model.
\newblock In \emph{Proceedings of the Twenty-Fourth AAAI Conference on Artificial Intelligence}, AAAI'10, page 1007–1012. AAAI Press.

\bibitem[{Churchill and Singh(2022)}]{TMevolution}
Rob Churchill and Lisa Singh. 2022.
\newblock \href {https://doi.org/10.1145/3507900} {The evolution of topic modeling}.
\newblock \emph{ACM Comput. Surv.}, 54(10s).

\bibitem[{Dasgupta and Hsu(2008)}]{flatTopic}
Sanjoy Dasgupta and Daniel Hsu. 2008.
\newblock \href {https://doi.org/10.1145/1390156.1390183} {Hierarchical sampling for active learning}.
\newblock In \emph{Proceedings of the 25th International Conference on Machine Learning}, ICML '08, page 208–215, New York, NY, USA. Association for Computing Machinery.

\bibitem[{Dieng et~al.(2020)Dieng, Ruiz, and Blei}]{ETMtopic}
Adji~B. Dieng, Francisco J.~R. Ruiz, and David~M. Blei. 2020.
\newblock \href {https://doi.org/10.1162/tacl_a_00325} {Topic modeling in embedding spaces}.
\newblock \emph{Transactions of the Association for Computational Linguistics}, 8:439--453.

\bibitem[{Doan and Hoang(2021)}]{BenchmarkingNT}
Thanh-Nam Doan and Tuan-Anh Hoang. 2021.
\newblock Benchmarking neural topic models: An empirical study.
\newblock In \emph{Findings}.

\bibitem[{Fang et~al.(2023)Fang, Alqazlan, Liu, He, and Procter}]{fang2023usercentered}
Zheng Fang, Lama Alqazlan, Du~Liu, Yulan He, and Rob Procter. 2023.
\newblock \href {http://arxiv.org/abs/2304.01774} {A user-centered, interactive, human-in-the-loop topic modelling system}.

\bibitem[{Fisher(1935)}]{ANOVA}
Ronald~A. Fisher. 1935.
\newblock \emph{The Design of Experiments}.
\newblock Oliver and Boyd, Edinburgh.

\bibitem[{Glaser and Strauss(2017)}]{glaser2017discovery}
Barney Glaser and Anselm Strauss. 2017.
\newblock \emph{Discovery of grounded theory: Strategies for qualitative research}.
\newblock Routledge.

\bibitem[{Grootendorst(2022)}]{bertopicMark}
Maarten Grootendorst. 2022.
\newblock \href {http://arxiv.org/abs/2203.05794} {Bertopic: Neural topic modeling with a class-based tf-idf procedure}.

\bibitem[{Hoyle et~al.(2021)Hoyle, Goel, Hian-Cheong, Peskov, Boyd-Graber, and Resnik}]{hoyle2021automated}
Alexander Hoyle, Pranav Goel, Andrew Hian-Cheong, Denis Peskov, Jordan Boyd-Graber, and Philip Resnik. 2021.
\newblock Is automated topic model evaluation broken? the incoherence of coherence.
\newblock In \emph{Advances in Neural Information Processing Systems}, volume~34, pages 2018--2033.

\bibitem[{Hoyle et~al.(2022)Hoyle, Goel, Sarkar, and Resnik}]{neuralbroken}
Alexander~Miserlis Hoyle, Pranav Goel, Rupak Sarkar, and Philip Resnik. 2022.
\newblock \href {https://doi.org/10.18653/v1/2022.findings-emnlp.390} {Are neural topic models broken?}
\newblock In \emph{Findings of the Association for Computational Linguistics: EMNLP 2022}, pages 5321--5344, Abu Dhabi, United Arab Emirates. Association for Computational Linguistics.

\bibitem[{Kleinberg(2002)}]{NIPS2002_43e4e6a6}
Jon Kleinberg. 2002.
\newblock \href {https://proceedings.neurips.cc/paper_files/paper/2002/file/43e4e6a6f341e00671e123714de019a8-Paper.pdf} {An impossibility theorem for clustering}.
\newblock In \emph{Advances in Neural Information Processing Systems}, volume~15. MIT Press.

\bibitem[{Krommyda et~al.(2021)Krommyda, Rigos, Bouklas, and Amditis}]{informatics8010019}
Maria Krommyda, Anastasios Rigos, Kostas Bouklas, and Angelos Amditis. 2021.
\newblock \href {https://doi.org/10.3390/informatics8010019} {An experimental analysis of data annotation methodologies for emotion detection in short text posted on social media}.
\newblock \emph{Informatics}, 8(1).

\bibitem[{Lee(2022)}]{tomotopy}
Minchul Lee. 2022.
\newblock \href {https://doi.org/10.5281/zenodo.6868418} {bab2min/tomotopy: 0.12.3}.

\bibitem[{Li et~al.(2024)Li, Mondal, Liang, Nghiem, and Boyd-Graber}]{li2024cfmatch}
Zongxia Li, Ishani Mondal, Yijun Liang, Huy Nghiem, and Jordan~Lee Boyd-Graber. 2024.
\newblock \href {http://arxiv.org/abs/2402.11161} {Panda (pedantic answer-correctness determination and adjudication):improving automatic evaluation for question answering and text generation}.

\bibitem[{Lin(2009)}]{lin_bmcbioinformatics2009}
Jimmy Lin. 2009.
\newblock Is searching full text more effective than searching abstracts?
\newblock \emph{BMC Bioinformatics}, 10:46.

\bibitem[{Lindstedt(2019)}]{Lindstedt2019}
Nathan~C. Lindstedt. 2019.
\newblock \href {https://doi.org/10.1177/2329496519846505} {Structural topic modeling for social scientists: A brief case study with social movement studies literature, 2005–2017}.
\newblock \emph{Social Currents}, 6(4):307--318.

\bibitem[{Mcauliffe and Blei(2007)}]{slda}
Jon Mcauliffe and David Blei. 2007.
\newblock Supervised topic models.
\newblock \emph{Advances in neural information processing systems}, 20.

\bibitem[{McHugh(2012)}]{McHugh2012}
Mary~L McHugh. 2012.
\newblock Interrater reliability: the kappa statistic.
\newblock \emph{Biochemia Medica}, 22(3):276--282.

\bibitem[{McInnes et~al.(2017)McInnes, Healy, and Astels}]{McInnes2017hdbscanHD}
Leland McInnes, John Healy, and S.~Astels. 2017.
\newblock \href {https://api.semanticscholar.org/CorpusID:53231359} {hdbscan: Hierarchical density based clustering}.
\newblock \emph{J. Open Source Softw.}, 2:205.

\bibitem[{McInnes et~al.(2020)McInnes, Healy, and Melville}]{mcinnes2020umap}
Leland McInnes, John Healy, and James Melville. 2020.
\newblock \href {http://arxiv.org/abs/1802.03426} {Umap: Uniform manifold approximation and projection for dimension reduction}.

\bibitem[{Mikolov et~al.(2013)Mikolov, Chen, Corrado, and Dean}]{word2vec}
Tomas Mikolov, Kai Chen, Gregory~S. Corrado, and Jeffrey Dean. 2013.
\newblock \href {https://api.semanticscholar.org/CorpusID:5959482} {Efficient estimation of word representations in vector space}.
\newblock In \emph{International Conference on Learning Representations}.

\bibitem[{Mitchell(1999)}]{misc_twenty_newsgroups_113}
Tom Mitchell. 1999.
\newblock {Twenty Newsgroups}.
\newblock UCI Machine Learning Repository.
\newblock {DOI}: https://doi.org/10.24432/C5C323.

\bibitem[{Muthukrishna et~al.(2019)Muthukrishna, Narayan, Mandel, Biswas, and Hložek}]{Muthukrishna_2019}
Daniel Muthukrishna, Gautham Narayan, Kaisey~S. Mandel, Rahul Biswas, and Renée Hložek. 2019.
\newblock \href {https://doi.org/10.1088/1538-3873/ab1609} {Rapid: Early classification of explosive transients using deep learning}.
\newblock \emph{Publications of the Astronomical Society of the Pacific}, 131(1005):118002.

\bibitem[{Papadia et~al.(2023)Papadia, Pacella, Perrone, and Giliberti}]{ItalianCaseStudy}
Gabriele Papadia, Massimo Pacella, Massimiliano Perrone, and Vincenzo Giliberti. 2023.
\newblock \href {https://doi.org/10.3390/a16020094} {A comparison of different topic modeling methods through a real case study of italian customer care}.
\newblock \emph{Algorithms}, 16(2).

\bibitem[{Paul and Dredze(2011)}]{Paul2011YouAW}
Michael~J. Paul and Mark Dredze. 2011.
\newblock \href {https://api.semanticscholar.org/CorpusID:9270435} {You are what you tweet: Analyzing twitter for public health}.
\newblock \emph{Proceedings of the International AAAI Conference on Web and Social Media}.

\bibitem[{Phan et~al.(2008)Phan, Nguyen, and Horiguchi}]{classification}
Xuan~Hieu Phan, Minh~Le Nguyen, and Susumu Horiguchi. 2008.
\newblock \href {https://api.semanticscholar.org/CorpusID:16198890} {Learning to classify short and sparse text \& web with hidden topics from large-scale data collections}.
\newblock In \emph{The Web Conference}.

\bibitem[{Poursabzi-Sangdeh et~al.(2016)Poursabzi-Sangdeh, Boyd-Graber, Findlater, and Seppi}]{alto}
Forough Poursabzi-Sangdeh, Jordan Boyd-Graber, Leah Findlater, and Kevin Seppi. 2016.
\newblock \href {https://doi.org/10.18653/v1/P16-1110} {{ALTO}: Active learning with topic overviews for speeding label induction and document labeling}.
\newblock In \emph{Proceedings of the 54th Annual Meeting of the Association for Computational Linguistics (Volume 1: Long Papers)}, pages 1158--1169, Berlin, Germany. Association for Computational Linguistics.

\bibitem[{Raeburn(2022)}]{Raeburn2022ContextSwitching}
Alicia Raeburn. 2022.
\newblock {Context switching is killing your productivity}.
\newblock \url{https://asana.com/resources/context-switching}.
\newblock Accessed: insert access date here.

\bibitem[{Rand(1971)}]{RandIndex}
William~M Rand. 1971.
\newblock Objective criteria for the evaluation of clustering methods.
\newblock \emph{Journal of the American Statistical association}, 66(336):846--850.

\bibitem[{Reimers and Gurevych(2019)}]{Reimers2019SentenceBERTSE}
Nils Reimers and Iryna Gurevych. 2019.
\newblock \href {https://api.semanticscholar.org/CorpusID:201646309} {Sentence-bert: Sentence embeddings using siamese bert-networks}.
\newblock In \emph{Conference on Empirical Methods in Natural Language Processing}.

\bibitem[{R\"{o}der et~al.(2015)R\"{o}der, Both, and Hinneburg}]{coherence}
Michael R\"{o}der, Andreas Both, and Alexander Hinneburg. 2015.
\newblock \href {https://doi.org/10.1145/2684822.2685324} {Exploring the space of topic coherence measures}.
\newblock WSDM '15, page 399–408, New York, NY, USA. Association for Computing Machinery.

\bibitem[{Rosenblatt(1958)}]{incrementallearning}
Frank Rosenblatt. 1958.
\newblock The perceptron: a probabilistic model for information storage and organization in the brain.
\newblock \emph{Psychological review}, 65(6):386.

\bibitem[{Settles(2012)}]{active-learning}
Burr Settles. 2012.
\newblock Active learning (synthesis lectures on artificial intelligence and machine learning).
\newblock In \emph{Findings}.

\bibitem[{Shannon(1948)}]{Shannon1948}
Claude~E. Shannon. 1948.
\newblock A mathematical theory of communication.
\newblock \emph{Bell System Technical Journal}, 27:379--423, 623--656.

\bibitem[{Strehl and Ghosh(2003)}]{NMI}
Alexander Strehl and Joydeep Ghosh. 2003.
\newblock \href {https://doi.org/10.1162/153244303321897735} {Cluster ensembles --- a knowledge reuse framework for combining multiple partitions}.
\newblock \emph{J. Mach. Learn. Res.}, 3(null):583–617.

\bibitem[{Sundqvist et~al.(2022)Sundqvist, Chiquet, and Rigaill}]{ARI}
Martina Sundqvist, Julien Chiquet, and Guillem Rigaill. 2022.
\newblock \href {https://doi.org/10.1007/s00180-022-01230-7} {Adjusting the adjusted rand index: A multinomial story}.
\newblock \emph{Comput. Stat.}, 38(1):327–347.

\bibitem[{Tang et~al.(2019)Tang, Li, and Jin}]{tang-etal-2019-topic}
Hongyin Tang, Miao Li, and Beihong Jin. 2019.
\newblock \href {https://doi.org/10.18653/v1/D19-1513} {A topic augmented text generation model: Joint learning of semantics and structural features}.
\newblock In \emph{Proceedings of the 2019 Conference on Empirical Methods in Natural Language Processing and the 9th International Joint Conference on Natural Language Processing (EMNLP-IJCNLP)}, pages 5090--5099, Hong Kong, China. Association for Computational Linguistics.

\bibitem[{{U.S. Department of Commerce}(2020)}]{NISTSP1190GB16}
{U.S. Department of Commerce}. 2020.
\newblock \href {https://doi.org/10.6028/NIST.SP.1190GB-16} {{Community Resilience Planning Guide for Buildings and Infrastructure Systems}}.
\newblock Technical Report NIST SP 1190GB-16, National Institute of Standards and Technology.

\bibitem[{Wang et~al.(2019)Wang, Gan, Xu, Zhang, Wang, Shen, Chen, and Carin}]{wang-etal-2019-topic}
Wenlin Wang, Zhe Gan, Hongteng Xu, Ruiyi Zhang, Guoyin Wang, Dinghan Shen, Changyou Chen, and Lawrence Carin. 2019.
\newblock \href {https://doi.org/10.18653/v1/N19-1015} {Topic-guided variational auto-encoder for text generation}.
\newblock In \emph{Proceedings of the 2019 Conference of the North {A}merican Chapter of the Association for Computational Linguistics: Human Language Technologies, Volume 1 (Long and Short Papers)}, pages 166--177, Minneapolis, Minnesota. Association for Computational Linguistics.

\bibitem[{Wang et~al.(2020)Wang, Duan, Zhang, Wang, Tian, Chen, and Zhou}]{wang-etal-2020-friendly}
Zhengjue Wang, Zhibin Duan, Hao Zhang, Chaojie Wang, Long Tian, Bo~Chen, and Mingyuan Zhou. 2020.
\newblock \href {https://doi.org/10.18653/v1/2020.emnlp-main.35} {Friendly topic assistant for transformer based abstractive summarization}.
\newblock In \emph{Proceedings of the 2020 Conference on Empirical Methods in Natural Language Processing (EMNLP)}, pages 485--497, Online. Association for Computational Linguistics.

\bibitem[{Wilson and Chew(2010)}]{termWeight}
A.~T. Wilson and P.~A. Chew. 2010.
\newblock Term weighting schemes for latent dirichlet allocation.
\newblock In \emph{Human Language Technologies: The 2010 Annual Conference of the North American Chapter of the Association for Computational Linguistics}, pages 465--473. Association for Computational Linguistics.

\bibitem[{Xu and Eguchi(2022)}]{xu2022supervised}
W~Xu and K~Eguchi. 2022.
\newblock \href {https://doi.org/10.1371/journal.pone.0277104} {A supervised topic embedding model and its application}.
\newblock \emph{PLoS One}, 17(11):e0277104.
\newblock PMID: 36331905; PMCID: PMC9635756.

\bibitem[{Zhang et~al.(2019)Zhang, Ding, and Song}]{zhang2019sp10k}
Hongming Zhang, Hantian Ding, and Yangqiu Song. 2019.
\newblock \href {http://arxiv.org/abs/1906.02123} {Sp-10k: A large-scale evaluation set for selectional preference acquisition}.

\bibitem[{Zhao et~al.(2021)Zhao, Phung, Huynh, Jin, Du, and Buntine}]{ijcai2021p638}
He~Zhao, Dinh Phung, Viet Huynh, Yuan Jin, Lan Du, and Wray Buntine. 2021.
\newblock \href {https://doi.org/10.24963/ijcai.2021/638} {Topic modelling meets deep neural networks: A survey}.
\newblock In \emph{Proceedings of the Thirtieth International Joint Conference on Artificial Intelligence, {IJCAI-21}}, pages 4713--4720. International Joint Conferences on Artificial Intelligence Organization.
\newblock Survey Track.

\bibitem[{Zhao et~al.(2020)Zhao, Rai, Du, Buntine, Phung, and Zhou}]{pmlr-v108-zhao20c}
He~Zhao, Piyush Rai, Lan Du, Wray Buntine, Dinh Phung, and Mingyuan Zhou. 2020.
\newblock \href {https://proceedings.mlr.press/v108/zhao20c.html} {Variational autoencoders for sparse and overdispersed discrete data}.
\newblock In \emph{Proceedings of the Twenty Third International Conference on Artificial Intelligence and Statistics}, volume 108 of \emph{Proceedings of Machine Learning Research}, pages 1684--1694. PMLR.

\bibitem[{Zhao et~al.(2011)Zhao, Jiang, Weng, He, Lim, Yan, and Li}]{Zhao2011ComparingTA}
Wayne~Xin Zhao, Jing Jiang, Jianshu Weng, Jing He, Ee-Peng Lim, Hongfei Yan, and Xiaoming Li. 2011.
\newblock \href {https://api.semanticscholar.org/CorpusID:7865163} {Comparing twitter and traditional media using topic models}.
\newblock In \emph{European Conference on Information Retrieval}.

\bibitem[{Zhao(2005)}]{purity}
Ying Zhao. 2005.
\newblock \emph{Criterion Functions for Document Clustering}.
\newblock Ph.D. thesis, USA.
\newblock AAI3180039.

\bibitem[{Zhou et~al.(2024)Zhou, Xu, Liu, An, Ai, and Huang}]{zhou2024explore}
Yuhang Zhou, Paiheng Xu, Xiaoyu Liu, Bang An, Wei Ai, and Furong Huang. 2024.
\newblock \href {http://arxiv.org/abs/2311.08648} {Explore spurious correlations at the concept level in language models for text classification}.

\end{thebibliography}
\bibliographystyle{style/acl_natbib}

\section{Appendix} \label{sec:sections/200-appendix} \appendix
\section{Clustering Evaluation Metric Details}
\label{sec:metrics_details}
We list and show the calculation details of automated evaluation metrics discussed in Section~\ref{sec:metrics} for easy of reproducing our work in this section. Suppose the classifier is trained on existing documents with user input labels (5\% of the documents), and the classifier predicts labels for all the documents, and they are partitioned into clusters denoted as $\Omega = \{\omega_1, \omega_2, \ldots, \omega_K\}$. The official gold clusters are denoted as $C = \{c_1, c_2, \ldots, c_J\}$.

\paragraph{Purity} It is calculated by assigning each cluster to the class which is most frequent in the cluster, and counting the correctly assigned points in that cluster. The formula to calculate the purity between the predicted and the gold clusters is:
\begin{equation}
    \text{Purity}(\Omega, C) = \frac{1}{N} \sum_{k} \max_j | \omega_k \cap c_j |.
\end{equation}

$N$ is the total number of points, $\omega_k$ is the $kth$ cluster, $c_j$ is the $jth$ class. $\omega_k \cap c_j$ is the number of points in cluster $\omega_k$ that belongs to class $c_j$, and $\max_j$ is maximum number of class $c_j$ intersection with cluster $\omega_k$~\cite{purity}.

\paragraph{Adjusted Normalized Mutual Information} The Adjusted Normalized Mutual Information (\abr{anmi}) is an improved version of the Normalized Mutual Information (\abr{nmi}) metric used for comparing the similarity between two clusterings that adjusts for chance to make the score more robust and comparable across different situations:
\begin{equation}
    \text{ANMI} = \frac{2 \times (\abr{mi} - \abr{e}[\abr{mi}])}{(H\abr{(c)} + H\abr{(k)}) - 2 \times \abr{e}[\abr{mi}]}.
\end{equation}

The mutual information (\abr{mi}) measures how much information we know about the gold clustering by knowing about the predicted clustering. The expected mutual information \abr{e[mi]} is calculation of what the \abr{mi} would be if the predicted clusters were completely at random, but still considering the size of the clusters. H\abr{(k)} measures the randomness or disorder within the gold clustering and H\abr{(c)} measures the randomness or disorder within the predicted clustering--entropy. A higher entropy means higher randomness for the clusters~\cite{ANMI}.

\paragraph{Adjusted Rand Index} Rand Index (\abr{ri}) computes the similarity between two clustering by considering pairs that are assigned in the same or different clusters in the predicted and true clustering~\cite{RandIndex}. The formula for \abr{ri} is:
\begin{equation}
    RI = \frac{\abr{tp} + \abr{tn}}{\abr{tp} + \abr{tn} + \abr{fp} + \abr{fn}}.
\end{equation}

\abr{tp} is the number of pairs that are in the same set in both the predicted and gold clusters, and \abr{tn} is the number of pairs that are in different sets in the predicted and gold clusters. Otherwise, the pairs are either \abr{fp} or \abr{fn}. 

The Adjusted Rand Index (\abr{ari}) is the corrected-for-chance version of the \abr{ri}. It accounts for the fact that the RI score will increase as the number of clusters increases, even if the clustering is random:
\begin{equation}
    RI = \frac{\abr{ri} - \text{Expected \abr{ri}}}{\text{Max \abr{ri}} - \text{Expected \abr{ri}}}.
\end{equation}

Expected \abr{ri} is the expected value of the \abr{ri} under random labeling, respecting the marginal distributions of cluster sizes. Max \abr{ri} is the highest possible value that the \abr{ri} could take, given the constraints of the clustering problem. A Max \abr{ri} of 1.0 indicates two clusterings are identical, but when adjusting it for chance, Max \abr{ri} can be less than 1 depending on the distribution of cluster sizes.

\paragraph{Normalized Pointwise Mutual Information} \abr{npmi} evaluates how semantically related the top words in each topic are to the documents in that topic, which in turn reflects the quality of the generated topics by a topic model:
\begin{equation}
    \text{NPMI}(x, y) = \frac{\log \frac{P(x, y)}{P(x) \cdot P(y)}}{-\log P(x, y)}.
\end{equation}
P(x, y) represents the probability of words x and y co-occuring together in a set of documents, where P(x) and P(u) are probabilities of observing words x and y independently in the set of documents. 

\section{Study Group Details}
\label{sec:study_group_details}
We provide more details of implementation about our 6 study groups introduced in Section~\ref{sec:groups} with two components-- user experience and classifier training. 

\subsection{User Interface Experience}
\begin{figure*}[t]
    \centering
    \includegraphics[scale=1.10]{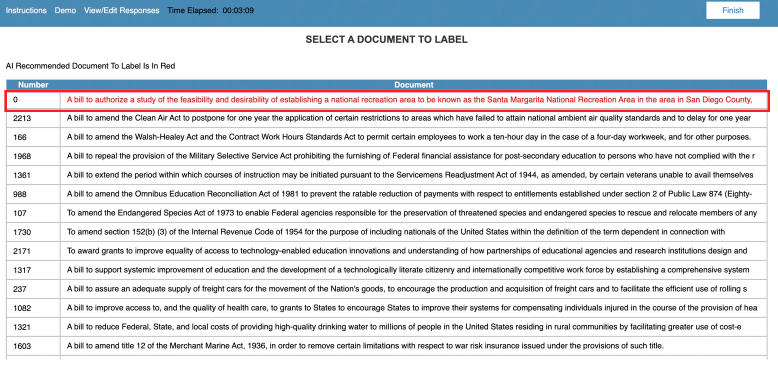}
    \caption{This is the overview (1) \abr{none} group. Users are not presented with topic overview, but active learning classifier picks the document based on the preference function and place it on top of the page.}
    \label{fig:notopic_interface1}
\end{figure*}

The baseline group (1) \abr{none} users only has access to a list of documents in the initial page shown in Figure~\ref{fig:notopic_interface1}. Active learning picks the most informative document and place it on top of the page so users can quickly selects it. Groups (2)-(6) with topic models have access to both active learning and topic overview shown in Figure~\ref{fig:interface1}. Users can explore the overall themes of the document sets then start labeling documents. After a user selects a document, topic model group users have access to the most related topics for the document, keywords, and highlighted keywords that are above 0.05 threshold for a selected topic shown in Figure~\ref{fig:interface2}. Group (1) \abr{none} users do not have access to the topic keywords and highlighted texts, but still retain the active learning basic features- the top three most relevant labels of the document predicted by the classifier. In all groups, users can click $\textit{submit \& next}$ button to automatically go to the next document selected by active learning or they can go back to the list of documents to select other documents.

\subsection{Classifier Training} 
(1) \abr{none} group users has a logistic regression classifier trained with their labeled documents. The classifier picks the next document based on the preference function with only \textit{tf-idf} as its input features. For group (2)-(6), we first compute the topic model probability features, where each document has an associated vector that contains probabilities it belongs to each topic. We encode the raw text features using \textit{tf-idf} first, and concatenate the topic vector with each encoded document features and train a classifier with user labeled documents. Classifiers in Group (2)-(6) have additional features generated by different topic models that can help classification to generalize better to unseen documents. Different topic models generate different features that can have diverse performance in downstream classification tasks.

\section{Simulated Experiment Details}\label{app:synthetic_experiment}
\paragraph{Training Topic Models}
We preprocess the dataset by tokenizing and filtering stopwords; we use a tf-idf threshold of three to remove rare and too-common words. 

For \abr{lda} and s\abr{lda}, we use the Tomotopy library ~\cite{tomotopy}, which uses Gibbs sampling to train classical topic models. To compare two datasets fairly, we chose $K=35$ topics for all five topic models in our group, which optimized average coherence. For \abr{lda} and s\abr{lda}, we use the term weight scheme ONE~\cite{termWeight}. s\abr{lda} takes more extra hyperparameters than \abr{lda} does. For s\abr{lda}, we also use binary-type response variables to indicate user input labels. Otherwise, \abr{lda} and s\abr{lda} use the default hyperparameter values. s\abr{lda} initially does not take in any response variables. We train \abr{lda} and s\abr{lda} with 2000 iterations until a smaller change of log-likelihood and NPMI coherence. 

For \abr{ctm}, we use SBERT \texttt{paraphrase-distilroberta-base-v2} to fetch sentence embeddings for the dataset, then concatenate them with BoW representation. We used \textit{CombinedTM}~\cite{kitty_ctm} with a 768 contextual size, with $K=35$ topics, and trained it with 250 epochs. We also use \texttt{paraphrase-distilroberta-base-v2} to fetch sentence ebmeddings to train Bertopic. For \abr{etm}, we use Word2Vec~\cite{word2vec} to encode documents and train it with 250 epochs.

\paragraph{Classifier Initialization and Features}
Since users are more likely to create more granular label specifications for each document. We used sub-labels as pseudo user-entered labels while using the more general labels as our gold standard. 

We use \texttt{sklearn} SGD as our classifier for active learning document selection.\footnote{We use hyperparameters: loss='log\_loss', penalty='l2', tolerance=10e-3, random\_state=42, learning\_rate='optimal', eta0=0.1, validation\_fraction=0.2, and alpha=0.000005.} We transform our raw dataset using unigram tf-idf as input features for the classifier. For \abr{lda}, s\abr{lda}, and \abr{ctm} groups, we also concatenate topic probability distributions for all the documents with unigram tf-idf features that also encode topic information to the classifier. Since the classifier requires at least two classes to be fitted, we pick random documents, and use sub-labels as surrogate user input labels, and activate the preference function until the classifier has at least two class labels. We use incremental leaning~\cite{incrementallearning} to fit and update the classifier, retaining originally learned parameters.\footnote{There are two exceptions we reinitialize the classifier: if a new label class is introduced to the classifier, we reinitialize the classifier and train it with labeled documents; if s\abr{lda} is updated with surrogate response variables, we rebuild the features by concatenating tf-idf features with new topic probability distributions, and restart the classifier with new features.} The classifier's predictions with the more general labels assess the clustering quality.



\paragraph{Simulated Experiment}
Upon analyzing the document lengths in our dataset, we deduced that considering individual reading speed variances, a user can feasibly label between 90 to 400 documents within an hour. For our simulated user study, we automatically run our algorithm for each group to input labels for 400 documents, constantly updating the classifier for every document labeled, and s\abr{lda} for every 50 documents labeled. Each group underwent 15 iterations of the experiment. For consistency, we aggregated the results by taking the median value for each document in each group. 

\paragraph{Validity of Simulated Experiments}
Of all the methods, \abr{ctm} consistently does better on purity, \abr{ari}, and \abr{anmi}, which underscores the right choice of \abr{ntm} can generate topic probability features that do better on classification. Such features, rooted in pre-trained embeddings, are perceived by compact machine learning models as more intuitive than the generative topic probabilities yielded by classical models like \abr{lda} and s\abr{lda}. s\abr{lda} and \abr{etm}, on the other side, is worse than \abr{lda}, where \abr{lda} remains competitive against two other \abr{ntm}s. The classifier without topic information falls short behind the classifier with topic information except for \abr{etm}. 

Our simple simulated experiments serve as a reliable proxy, allowing us to expect similar trends when actual human labeling is in play and to track the evolution of classifier predictions as more documents are labeled over time. However, we acknowledge that relying solely on simulated evaluation metrics has limitations. The classifier does not consider using topic keywords and topic overviews to create labels. Other factors, including fatigue and loss of attention, might also affect the quality of labels created by real users. Such metrics also do not capture the complete essence of user preferences, especially concerning the keywords produced by topic models, the highlighted keywords, or the specific documents recommended by the preference function.

\section{Dataset Details}\label{app:dataset}
The Bills have over 400,000 bills spanning from 1947 to 2009, where each bill is meticulously labeled with primary and secondary topics, as detailed in a comprehensive codebook.\footnote{\url{https://comparativeagendas.s3.amazonaws.com/codebookfiles/Codebook_PAP_2019.pdf.}} The latest iteration of this dataset has seen its topics labeled by adept human coders, who were trained using the preceding dataset version. The inter-annotator agreement was observed to be an impressive 95\% for primary topics and 75\% for secondary ones. Such extensive and refined labeling, carried out by trained annotators over numerous years, assures the dataset's label quality. The 20newsgroup is a popular benchmark dataset that has 6 major labels and 20 sub-labels. We remove duplicate documents, documents that are shorter than 30 tokens, documents that contain sensitive topics, and documents that the general public is not familiar with the Bills and 20newsgroup dataset.  

\section{User Label Evaluations}\label{app:sanity_check}
We do a sanity check on the 800 randomly selected labeled documents, to ensure users are creating meaningful labels. Within each group, we sort the users based on the summation of purity, \abr{ari}, \abr{anmi} at the end of the 61st minute in ascending order. We take the middle 8 users and randomly pick 200 labeled documents from each group. We have two annotators manually judge the user labels based on the following two criteria: 1. Can the user label be considered equivalent or a subfield of the gold label (major label and sub label)? 2. Does the user label reflect the contents of the passage? If the annotator rates `yes' for criteria 1, criteria 2 will be skipped. Otherwise, the annotator will need to read the actual passage to judge the quality of the user labels. Among 800 labeled documents, we have 787 documents that satisfy at least one of the two criteria, which ensures most users are making meaningful labels and carefully conducting the study.

\section{User Interface}\label{app:user_interface}
\begin{figure*}[t]
    \centering
    \includegraphics[scale=0.55]{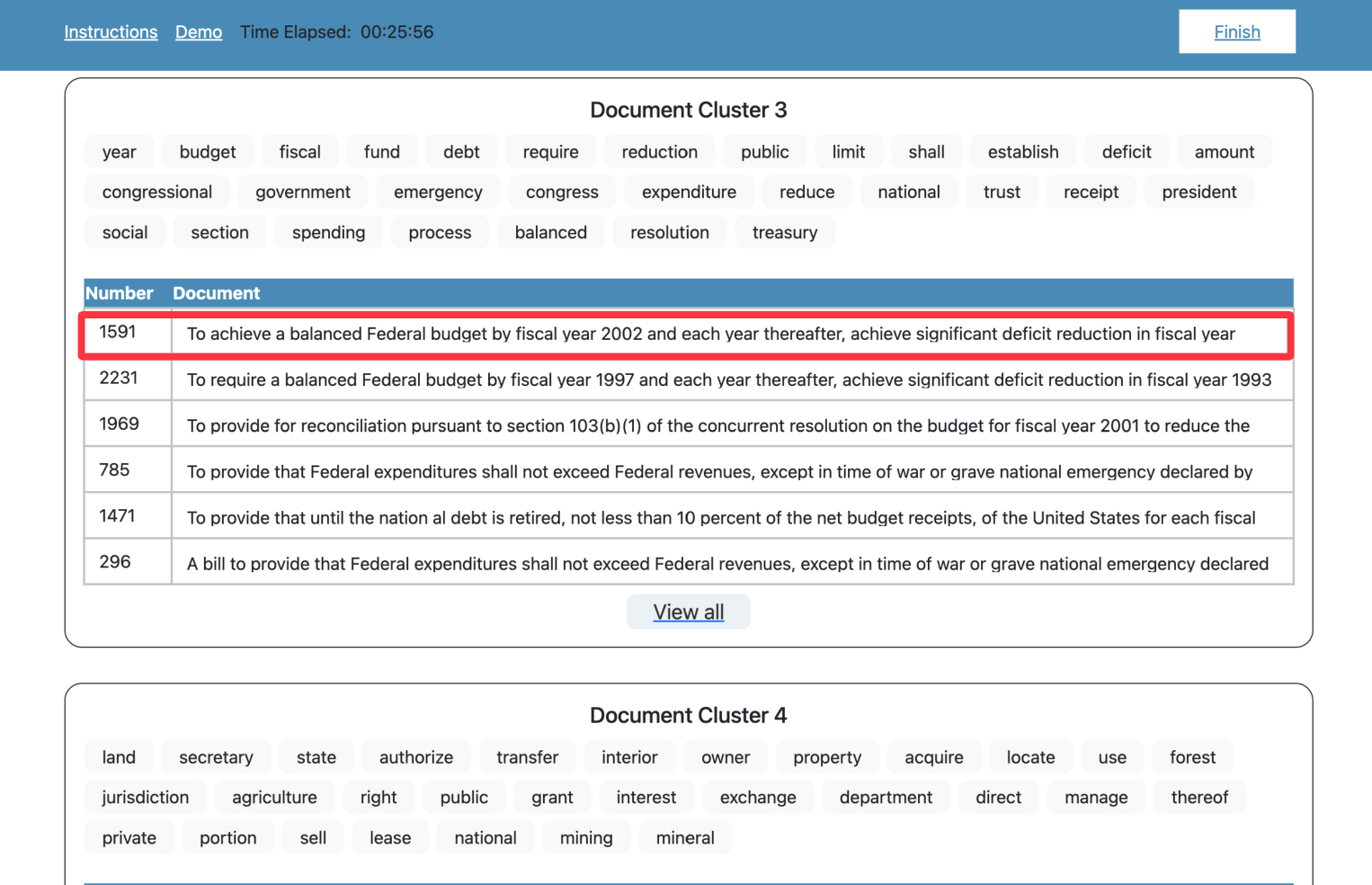}
    \caption{Under topic model settings, users are displayed all topics, keywords, and documents in each topic. If active learning picks a document, the topic and the document cluster containing that document will be displayed at the very top of this page. The document is also displayed on the top of the document cluster. For example, the document marked red is an example of a document picked by active learning. For the baseline, \abr{none} group, topic keywords, and document clusters are not displayed. All documents are displayed in one block, and the recommended document is always on top of the page above other documents.}
    \label{fig:interface1}
\end{figure*}

\begin{figure*}[t]
    \centering
    \includegraphics[scale=0.35]{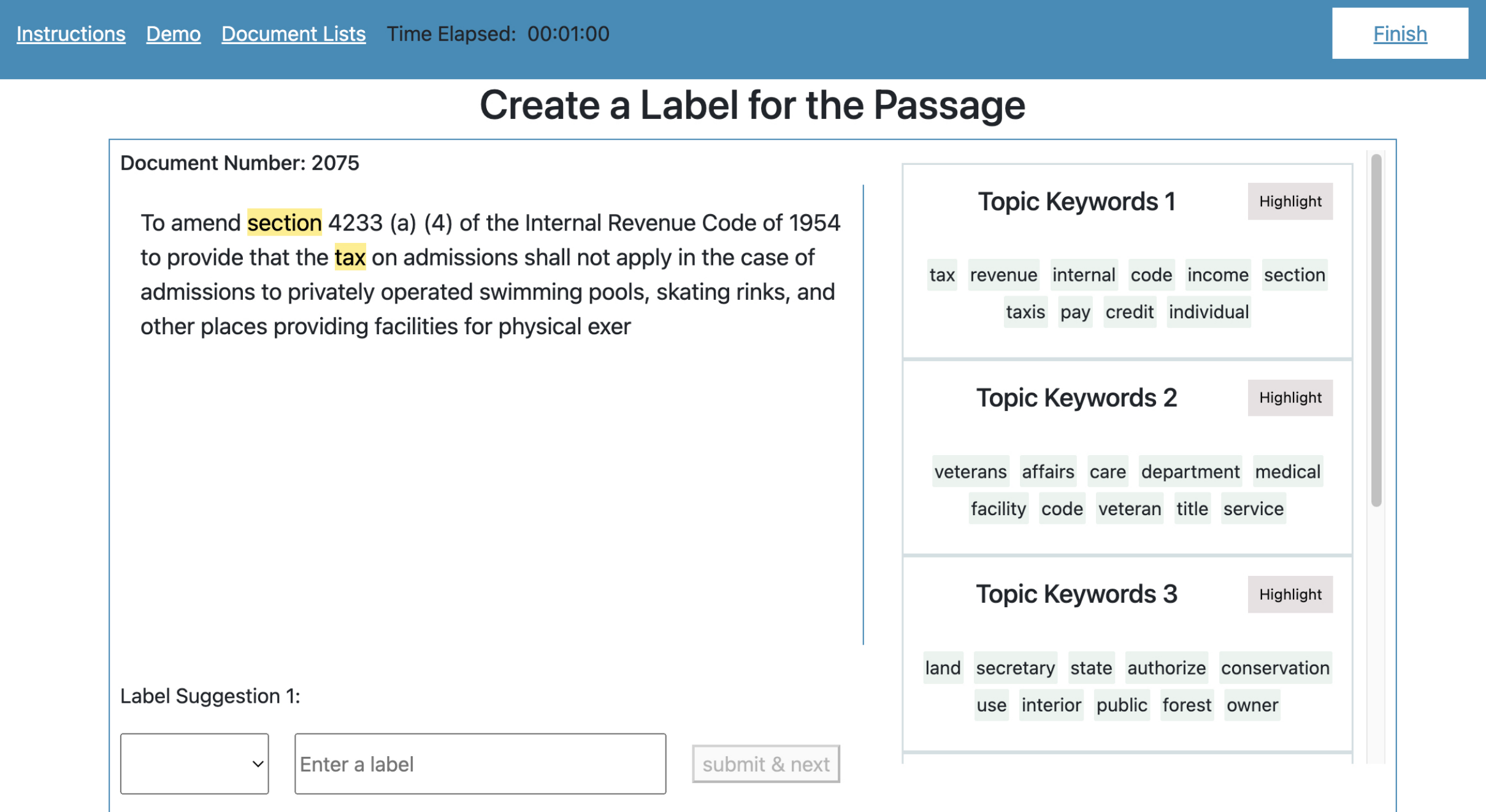}
    \caption{For a user-selected document, a user can either make a label for the document or skip the document. The top 5 most relevant topics and top keywords for the selected document are displayed on the right side. The highlight function helps users quickly find words that are above the 0.05 threshold for a chosen topic. Users could also select a label from the dropdown box, which the labels are ranked by softmax probabilities of the classifier, and the dropdown labels are what the users have created so far. For \abr{none}, the highlights and topics will not be available to the users.}
    \label{fig:interface2}
\end{figure*}

Figure~\ref{fig:interface1} and Figure~\ref{fig:interface2} show a basic layout of \abr{ctm} used in our user study. The keywords and document clusters will not be displayed to \abr{none} group users. Instead, a random list of documents are displayed to them in Figure~\ref{fig:interface1} page. In Figure~\ref{fig:interface2} page, \abr{none} users are not displayed with the \textit{Top Topic Keywords} and the highlighted texts.

\section{Topic Model Keywords}\label{app:keyword_tables}
Table~\ref{tab:topic_examples}, ~\ref{tab:mid_topic_examples}, and ~\ref{tab:low_topic_examples} show the 2 topics with highest, median, and lowest \abr{npmi} coherence scores for \abr{lda}, s\abr{lda}, \abr{ctm}, and s\abr{lda} trained with user input labels as response variables. The topic keywords generated by \abr{lda} are more general and inclusive while the topic keywords generated by \abr{ctm} are more specific and related to the top passages.





\begin{table*}[h]
\centering
\begin{tabularx}{\textwidth}{l l p{0.35\linewidth} X}
\hline
Model & NPMI & Keywords & Passage \\
\hline
\hline
\abr{lda} & 0.39 & exemption, income, dependent, increase, taxpayer, tax, spouse, personal, additional, include & To provide that certain survivor benefits received by a child under public retirement systems shall not be taken into account in determining whether the child is a dependent for income tax purposes. \\ 
\hline
\abr{lda} & 0.24 & tax, revenue, internal, code, income, section, taxis, pay, credit, individual & To amend the Internal Revenue Code of 1954 to include the sintering and burning of clay, shale, and slate used as lightweight aggregates as a treatment process considered as mining. \\ 
\hline
s\abr{lda} & 0.35 & rescind, control, authority, budget, president, special, impoundment, propose, transmit, section & To rescind certain budget authority proposed to be rescinded (R92-66) in a special message transmitted to the Congress by the President on March 20, 1992. \\ 
\hline
s\abr{lda} & 0.22 & tax, revenue, income, internal, code, exemption, section, individual, taxis, shall & To amend the Internal Revenue Code to provide that gain or loss from the sale or exchange of certain real estate shall be treated as a capital gain or loss. \\ 
\hline
\abr{ctm} & 0.50 & president, authority, propose, rescind, congress, special, impoundment, march, accordance, trasmit, message & A bill to rescind certain budget authority contained in the message of the President of January 27, 1978 (H. Doc. 95-285), transmitted pursuant to the Impoundment Control Act of 1974. \\ 
\hline
\abr{ctm} & 0.38 & exemption, include, taxpayer, personal,  additional, increase, dependent, spouse, income, old & To increase from $ \text{\$} $600 to $ \text{\$} $750 the personal income tax exemptions of a taxpayer (including the exemption for a spouse, the exemption for a dependent, and the additional exemption for old age, or blindness). \\ 
\hline
s\abr{lda}(user) & 0.42 & budget, rescind, control, president, authority, impoundment, congress, transmit, message, section & To amend part C of the Balanced Budget and Emergency Deficit Control Act of 1985 to extend the discretionary spending limits and pay-as-you-go through fiscal year 2009. \\ 
\hline
s\abr{lda}(user) & 0.26 & education, school, student, loan, program, secondary, institution, elementary, educational, teacher & To amend the Higher Education Act of 1965 to expand the loan forgiveness and loan cancellation programs for teachers, to provide loan forgiveness and loan cancellation programs for nurses, and for other purposes. \\ 
\hline
\end{tabularx}

\caption{Topic models automatically discover topics and themes in the Bills dataset. These topics give users a global sense of probable stories and themes in a dataset. We show the top 2 topics for each topic model and their relevant keywords and relevant passages. s\abr{lda} is the initial model without fitting with response variables, which is used for all users in s\abr{lda} group. s\abr{lda}(user) uses a pre-saved model, which is derived from the median calculations (median of summation of purity, \abr{ari}, \abr{anmi} among 15 users) across 15 users in s\abr{lda}. s\abr{lda}(user) generates top topics with higher top coherence scores than other models. The keywords also appear more often and are more related to passages.}
\label{tab:topic_examples}
\end{table*}

\begin{table*}[h]
\centering
\begin{tabularx}{\textwidth}{l l p{0.35\linewidth} X}
\hline
Model & NPMI & Keywords & Passage \\
\hline
\hline
\abr{lda} & 0.13 & water, wildlife, conservation, fish, establish, management, resource, national, development, coastal & To create a joint commission of the United States and the State of Alaska to make administrative determinations of navigability of inland nontidal waters in the State of Alaska for State selections. \\ 
\hline
\abr{lda} & 0.12 & food, drug, use, cosmetic, respect, human, child, information, intend, manufacturer & A bill to amend Sections 403 and 405 of the Federal Food, Drug, and Cosmetic Act to require that foods intended for human consumption be labeled to show the amount of sodium and potassium they contain. \\ 
\hline
s\abr{lda} & 0.10 & labor, section, employee, national, organization, fair, provision, relations, right, railway & To amend the Railroad Retirement Act of 1937 and the Social Security Act to eliminate those provisions which restrict the right of a spouse or survivor to receive benefits simultaneously under both acts. \\ 
\hline
s\abr{lda} & 0.07 & highway, title, section, amend, national, code, fund, system, construction, stat & A bill to supplement the Federal Aid Road Act, approved July 11, 1916, as amended and supplemented, to authorize appropriations for the construction of greatly needed rural local roads, and for other purposes. \\ 
\hline
\abr{ctm} & 0.07 & contract, standards, work, wage, contractor, cause, hour, fair, employer, employee & A bill to provide for the creditability of certain service in determining the order of retention for competing employees in a reduction in force affecting the Federal Grain Inspection Service. \\ 
\hline
abr{ctm} & 0.06 & revenue, internal, code, section, estate, sale, admission, value, treatment, relate & To amend section 112 (b) of the Internal Revenue Code (relating to recognition of gain in certain corporate liquidations) so that it will apply to cases where the transfer of all the property under the liquidation occurs within 1 calendar month in 1953. \\ 
\hline
s\abr{lda}(user) & 0.03 & program, establish, improve, development, system, promote, assist, provide, national, encourage & A bill to improve existing tertiary eye centers, to examine the delivery of eye care to the general public, and to study the feasibility of implementing a system of tertiary eye care centers throughout the United States. \\ 
\hline
s\abr{lda}(user) & 0.02 & state, fund, program, year, title, establish, assistance, construction, facility, authorize & To amend the National Housing Act to authorize the Secretary of Housing and Urban Development to insure mortgages for the acquisition, construction\dots \\ 
\hline
\end{tabularx}

\caption{The table shows the 18th and 19th coherent topics discovered by different topic models. The bottom 2 topics for s\abr{lda}(user) only have a few passages associated with each of them.}
\label{tab:mid_topic_examples}
\end{table*}

\begin{table*}[h]
\centering
\begin{tabularx}{\textwidth}{l l p{0.35\linewidth} X}
\hline
Model & NPMI & Keywords & Passage \\
\hline
\hline
\abr{lda} & -0.10 & person, foreign, prohibit, business, engage, country, trade, domestic, enable, stock & To provide an exception from certain group health plan requirements to allow small businesses to use pre-tax dollars to assist employees in the purchase of policies in the individual health insurance market, and for other purposes. \\ 
\hline
\abr{lda} & -0.08 & vessel, coast, guard, marine, specie, merchant, port, law, academy, endangered & To amend the Merchant Marine Act of 1936 and the Maritime Academy Act of 1958 to enlarge the mission of the U.S. Merchant Marine Academy and to assist in enlarging the mission of the State maritime academies. \\ 
\hline
s\abr{lda} & -0.12 & meat, product, inspection, state, continental, shelf, outer, poultry, import, land & A bill to modify the method of determining quantitative limitations on the importation of certain articles of meat and meat products, to apply quantitative limitations on the importation of certain additional articles of meat, meat products, and livestock, and for other purposes. \\ 
\hline
s\abr{lda} & -0.11 & fla, know, value, historic, shall, national, site, use, fort, dam & A bill to provide that the reservior formed by the lock and dam referred to as the Millers Ferry lock and dam on the Alabama River, Alabama, shall hereafter be known as the William Bill Dannelly Reservior. \\ 
\hline
\abr{ctm} & -0.29 & locate, convey, transfer, territory, memorial, historical, washington, smithsonian, city, conveyance & To provide for the conveyance of certain excess real property of the United States to the city of Mission, the city of McAllen, and the city of Edinburg, all situated in the State of Texas. \\ 
\hline
\abr{ctm} & -0.12 & highway, aid, interstate, road, alaska, system, fund, fla, commission, transportation & To amend section 5 of the Department of Transportation Act to authorize the National Transportation Safety Board to employ 5,000 investigators to carry out its powers and duties under that act. \\ 
\hline
s\abr{lda}(user) & -0.36 & gas, purpose, greenhouse, wheat, red, cheese, cheddar, operate, exist, standards & To provide that the rules of the Environmental Protection Agency entitled National Emission Standards for Hazardous Air Pollutants for Reciprocating Internal Combustion Engines\dots \\ 
\hline
s\abr{lda}(user) & -0.31 & gram, trans, drugs, deadline, intervention, temple, manatees, plains, ombudsman, leaseholder & To direct the Commissioner of Food and Drugs to revise the Federal regulations applicable to the declaration of the trans fat content of a food on the label and in the labeling of the food when such content is less than 0.5 gram. \\ 
\hline
\end{tabularx}

\caption{The table shows the least two coherent topics discovered by different topic models. The bottom 2 topics for s\abr{lda}(user) only have a few passages associated with each of them.}
\label{tab:low_topic_examples}
\end{table*}

 \label{appendix}

\end{document}